\documentclass[journal]{IEEEtran} 
\ifCLASSINFOpdf
\else
\fi
\usepackage{subcaption}                             
\usepackage{graphicx}           
\usepackage{amsmath} 
\usepackage{amssymb}
\usepackage{pgfplots}
\usepackage{pgfplotstable}
\usepackage{bm}
\usepackage{cite}
\usepackage{url}
\usepackage{makecell}                     
\usepackage{multirow} 
\usepackage{setspace}
\usepackage{cellspace}
\usepackage{booktabs}
\usepackage{enumitem} 
\usepackage{multirow}
\usepackage{breqn}
\usetikzlibrary{arrows,positioning} 
\usetikzlibrary{plotmarks}
\usetikzlibrary{calc}
\usepgfplotslibrary{groupplots}
\usetikzlibrary{patterns}
\usetikzlibrary{shapes.geometric}
\usetikzlibrary{fit}
\usetikzlibrary{arrows,positioning,shapes,backgrounds,fit} 

\usepackage{amssymb}
\usepackage{pifont}

\pgfplotsset{compat=1.18}
\usepackage{caption}     
\usepackage{svg}
\usepackage{hyperref} 
\usepackage{xurl}  
\usepackage{graphicx}            
\usepackage{amsmath} 
\usepackage{amssymb}
\usepackage{pgfplots}
\usepackage{pgfplotstable}
\usepackage{bm}
\usepackage{cite}
\usepackage{url}
\usepackage{makecell}
\usepackage{multirow}
\usepackage{setspace}
\usepackage{cellspace}
\usepackage{booktabs}
\usepackage{enumitem}
\usepackage{multirow}
\usepackage{breqn}
\usetikzlibrary{arrows,positioning}
\usetikzlibrary{plotmarks}
\usetikzlibrary{calc}
\usepgfplotslibrary{groupplots}
\usetikzlibrary{patterns}
\usetikzlibrary{shapes.geometric}
\usetikzlibrary{fit}
\usetikzlibrary{arrows,positioning,shapes,backgrounds,fit}

\pdfoutput=1

\begin{document}              
\title{Building Age Estimation: A New Multi-Modal Benchmark Dataset and Community Challenge}            
\author{Nikolaos Dionelis, Alessandra Feliciotti, Mattia Marconcini, Devis Peressutti, Nika Oman Kadunc, JaeWan Park, Hagai Raja Sinulingga, Steve Andreas Immanuel, Ba Tran, Caroline Arnold, Nicolas Longépé 
\thanks{N. Dionelis and N. Longépé are with the European Space Agency (ESA), $\Phi$-lab, ESRIN, Italy, e-mail: \{Nikolaos.Dionelis, Nicolas.Longepe\}@esa.int.}  
\thanks{A. Feliciotti and M. Marconcini are with MindEarth, Switzerland, e-mail: \{alessandra.feliciotti, mattia.marconcini\}@mindearth.ch.}
\thanks{D. Peressutti and N. O. Kadunc are with Sinergise/ Planet, Slovenia, email: \{devis.peressutti, nika.oman-kadunc\}@planet.com.}
\thanks{J. Park, H. R. Sinulingga, and S. A. Immanuel are with TelePIX, Seoul, South Korea, e-mail: \{eric.park, hagairaja, steve\}@telepix.net.}   
\thanks{B. Tran is with Axelspace Corporation, Tokyo, Japan, e-mail: thba1590@gmail.com.}   
\thanks{C. Arnold is with Helmholtz Institute Hereon, Germany and German Climate Computing Center DKRZ, Germany, e-mail: arnold@dkrz.de.}    
\thanks{Manuscript created October, 2024; received January, 2025; revised April 2025; revised July 2025. Previous version: \url{http://arxiv.org/pdf/2502.13818v2}.}}   

\markboth{Journal of \LaTeX\ Class Files,~Vol.~XX, No.~X, May~2025}%
{Shell \MakeLowercase{\textit{et al.}}: Bare Demo of IEEEtran.cls for IEEE Journals} 

\maketitle                                                                                           

\pdfoutput=1 

\begin{abstract}                               
Estimating the construction year of buildings is critical for advancing sustainability, as older structures often lack energy-efficient features. Sustainable urban planning relies on accurate building age data to reduce energy consumption and mitigate climate change. In this work, we introduce MapYourCity, a novel multi-modal benchmark dataset comprising top-view Very High Resolution (VHR) imagery, multi-spectral Earth Observation (EO) data from the Copernicus Sentinel-2 constellation, and co-localized street-view images across various European cities. Each building is labeled with its construction epoch, and the task is formulated as a seven-class classification problem covering periods from 1900 to the present. To advance research in EO generalization and multi-modal learning, we organized a community-driven data challenge in 2024, hosted by ESA $\Phi$-lab, which ran for four months and attracted wide participation.

This paper presents the Top-4 performing models from the challenge and their evaluation results. We assess model generalization on cities excluded from training to prevent data leakage, and evaluate performance under missing modality scenarios, particularly when street-view data is unavailable. Results demonstrate that building age estimation is both feasible and effective, even in previously unseen cities and when relying solely on top-view satellite imagery (i.e. with VHR and Sentinel-2 images). The MapYourCity dataset thus provides a valuable resource for developing scalable, real-world solutions in sustainable urban analytics.

\end{abstract} 

\IEEEpeerreviewmaketitle                   

\section{Introduction}       
\label{sec:intro}  
\noindent      
Estimation of the age of buildings in cities is important for sustainability, urban planning, and structural safety purposes.                  
Sustainable buildings minimize energy consumption and are a key part of responsible and sustainable urban planning and development that seeks to effectively combat climate change.   
The urgency to assess the energy efficiency of buildings has never been greater, as it plays a crucial role in achieving key sustainability goals.          
Building age plays a significant role in energy modelling and urban policy development. It can serve as a useful proxy for estimating energy efficiency. However, globally consistent and homogeneous data on the construction period of buildings is currently lacking. Earth Observation (EO) and geospatial analytics offer promising solutions to help build such a global database, enabling large-scale and consistent mapping of building characteristics. Since no single EO mission provides direct information on construction age, Artificial Intelligence (AI) can be leveraged to fuse and interpret heterogeneous datasets, allowing for indirect yet robust estimation of building age.      
To help fill this gap, we organized a data challenge in 2024, named MapYourCity, which includes the preparation and release of a multi-modal dataset and a competition designed to foster the use of the latest AI architectures for building age estimation. 

In this paper, we introduce a new benchmark dataset, consisting of three modalities with street-view ground images, top-view Very High Resolution (VHR) satellite images at $50$    cm resolution, and Copernicus Sentinel-2 (S-2) \textit{multi-spectral} satellite data.    
All the modalities are co-localized with respect to the specific building under study, as well as its label of the construction epoch.    
The novel multi-modal benchmark dataset is suitable for correlation learning between the different modalities, as well as for data fusion. We present the MapYourCity challenge organized around this dataset, evaluate the most performing models, and demonstrate the effectiveness of fusing these modalities for accurately predicting building age, a key factor for sustainability, energy efficiency, urban planning and development, and safety.  
The problem is formulated as a classification task with seven classes for the construction epoch of buildings, ranging from 1900 to nowadays. This paper evaluates the generalization performance of the models on cities that are \textit{not}   included in the training dataset, i.e. on new and previously unseen cities. 
We examine the performance of the models when both training and testing are performed with all three modalities, but also when testing does \textit{not} include street-view images (top-view VHR and S-2 images only).\begin{figure}[t]                                  
  \centering \includegraphics[height=7.5cm, width=8.7cm]{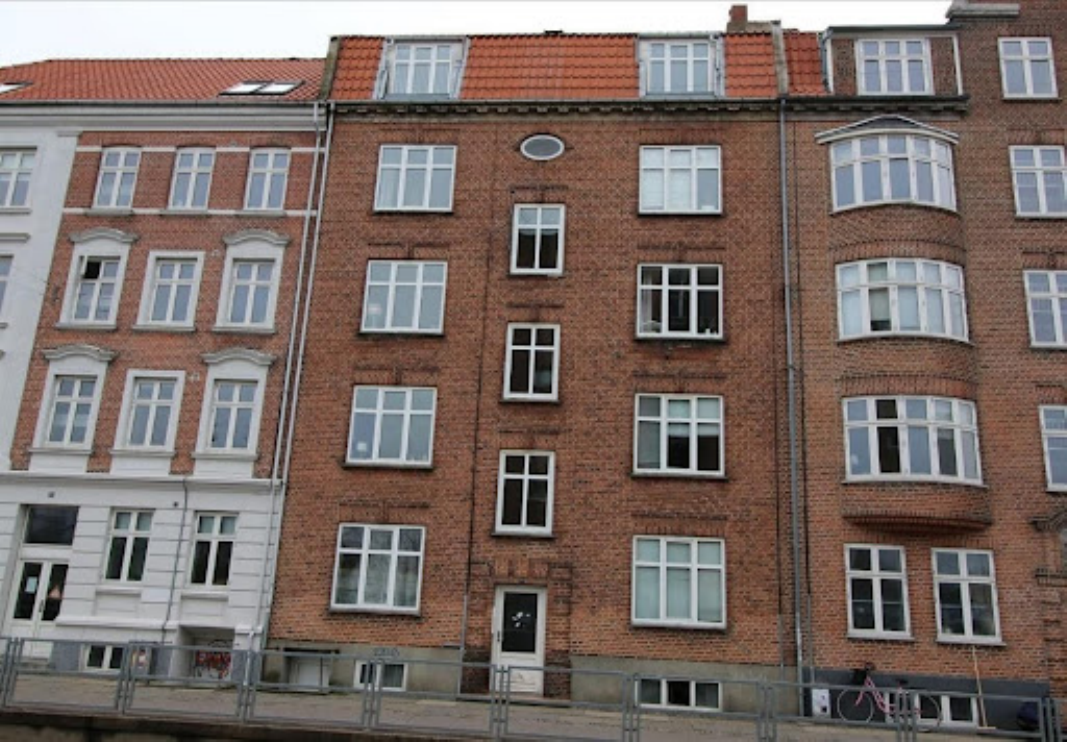} 
    \centering \includegraphics[height=7.5cm, width=8.8cm]{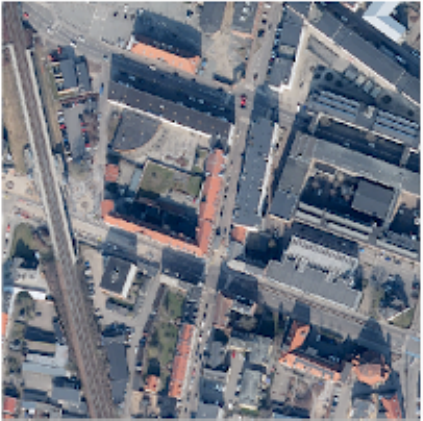} 
  \caption{Example of collocated street-view picture (Top) and top-view VHR image (Bottom).}      
  \label{fig:short1}                    
\end{figure}

\begin{figure}[t]                                   
   \centering \includegraphics[height=7.5cm]{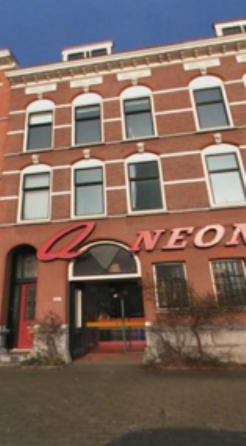}
    \centering \includegraphics[height=7.5cm, width=8.8cm]{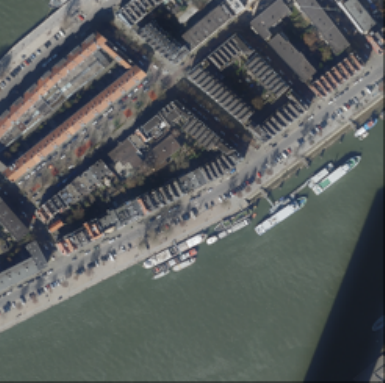}
  \caption{Example of collocated street-view picture (Top) and top-view VHR image (Bottom). The varying size of the street-view image corresponds to what is available and included in the MapYourCity dataset.}           
  \label{fig:short2}                    
\end{figure}

This paper is structured as follows. In Sec. \ref{sec:rel}, we review related work on combining street-view with EO imagery, highlighting recent advances in multi-modal learning and Transformer-based architectures. Next, Sec. \ref{sec:datatask}  introduces the novel multi-modal dataset, which includes co-localized street-view images, top-view VHR satellite imagery, and S-2 EO data, along with the associated challenge task focused on predicting the construction epoch of buildings across European cities. In Secs. \ref{sec:mainmodels}-\ref{sec:thibes}, we present the top four performing models from the challenge, detailing their data preprocessing strategies, architectural choices, and evaluation results. The baseline model provided in the starter toolkit for all challenge participants is described in  Sec. \ref{sec:baselinemo}. Sec. \ref{sec:discdiscussion} provides a comparative discussion of these models, emphasizing the main features and design decisions that contributed to their success. Finally, Sec. \ref{sec:conc} concludes the paper by summarizing the key findings and outlining future directions for research in sustainable urban analytics using multi-modal EO data.

\section{Related Work} \label{sec:rel}      
\noindent The value of street-view imagery in urban analytics has become increasingly evident in recent literature \cite{StreetView1, chen2024global}. It provides rich visual context for analyzing the built environment, vegetation, transportation networks, and urban infrastructure, supporting studies in health, socio-economic development, sustainability, and urban planning  \cite{StreetView1, streetpaper}. Street-view data also play a key role in assessing urban mobility, spatial data infrastructures, and urban greenery \cite{2024_global_streetscapes, StreetView1, streetpaper}. A notable example is the Global Streetscapes dataset introduced in \cite{2024_global_streetscapes}, which comprises over 10 million street-view images from $688$ cities around the Earth, from $212$ countries and different regions.  These images, crowdsourced from Mapillary and KartaView, are enriched with metadata including geo-location, acquisition time, and semantic, perceptual, and contextual information—making them a valuable resource for large-scale urban analysis.

Recent AI models increasingly explore data fusion between street-view and top-view satellite imagery. Due to the substantial differences in resolution, scale, object characteristics, and visual appearance between these modalities, late fusion (i.e. combining latent feature representations after independent encoding) often yields better performance than early data fusion approaches. Street-view images offer detailed ground-level perspectives, while top-view satellite data cover broader geographic areas from a nadir viewpoint. Effectively integrating these heterogeneous sources remains an open research challenge in multi-modal learning and remote sensing. The model proposed in \cite{Zhu1} maps building functions such as commercial, residential, public, and industrial, by jointly leveraging street-view and top-view imagery through a decision-level data fusion strategy. Separate models are trained for each modality, and their predictions are then combined to enhance performance and classification accuracy. The approach proposed in \cite{Zhu1} effectively integrates complementary information from both perspectives, enabling better and more precise functional mapping of urban buildings.\begin{figure}[t]                                  
  \centering \includegraphics[height=7.5cm, width=8.8cm]{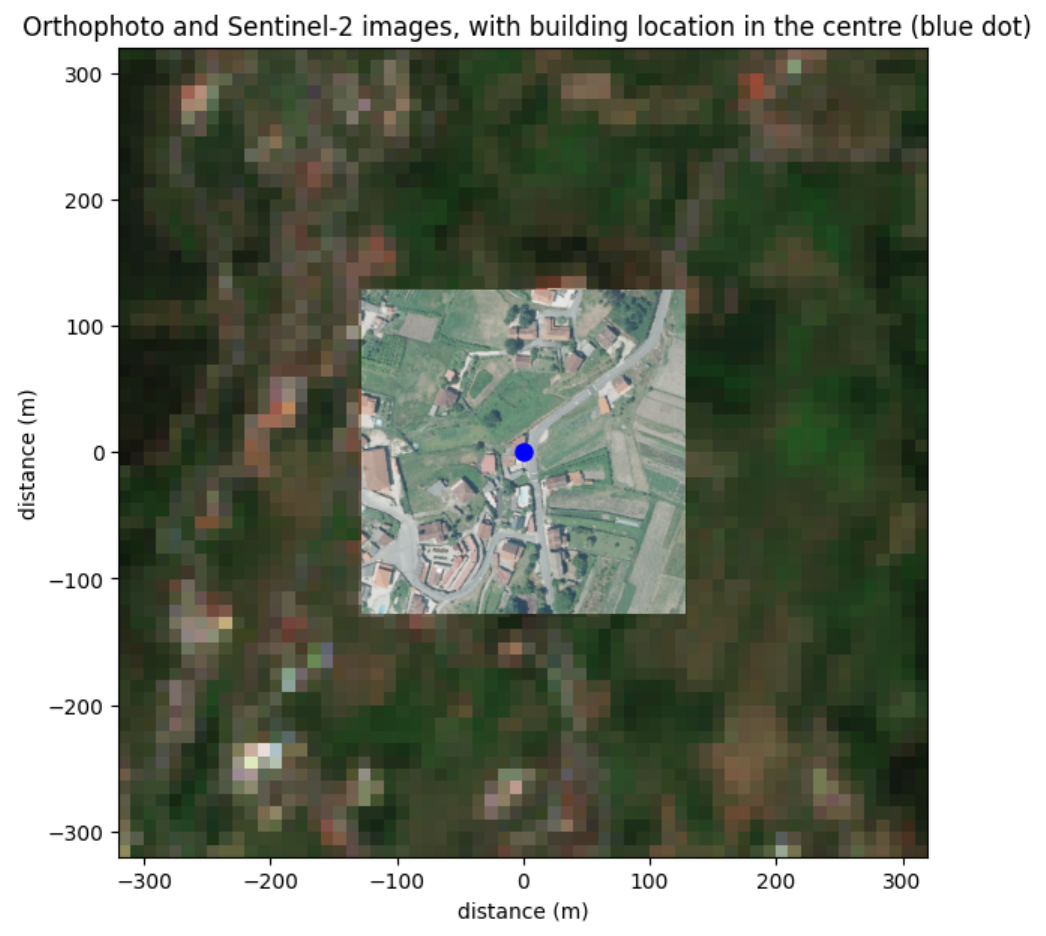}      
  \caption{Example top-view satellite VHR image and the corresponding geo-localized S-2 image (here, RGB bands only for visualization purposes).}        
  \label{fig:short1vv54}                       
\end{figure}

Transformer models have demonstrated superior performance over traditional architectures such as Convolutional Neural Networks (CNNs), U-Net, and Residual Networks (ResNet), particularly in tasks requiring global context modeling. In \cite{transf1}, the Transformer-based model TransGeo is introduced for geo-location matching between street-view and top-view aerial images. Leveraging the attention mechanism, TransGeo performs attention-guided non-uniform cropping, allowing the model to focus selectively on informative image regions while ignoring irrelevant areas. This capability to model long-range dependencies and adaptively attend to spatial features makes Transformers especially effective for complex multi-modal tasks in remote sensing and urban analytics. Recent Transformer-based vision models have proven highly effective, achieving strong performance across various tasks \cite{transf2}. For instance, the model proposed in \cite{transf2} applies semantic segmentation to unlabeled top-view aerial imagery using domain adaptation techniques, and is built upon the SegFormer architecture \cite{transf3}. The results of the model presented in \cite{transf2} have been used for \textit{cross-view} geo-location matching of street-view images and top-view aerial images in \cite{transf4}. Accurately aligning these two modalities remains a challenging task due to their distinct perspectives and different visual characteristics \cite{transf5}. In addition, street-view images are also used in \cite{newnewpaperstreet}.

Several datasets provide top-view satellite or aerial imagery for semantic segmentation tasks. One notable example is OpenEarthMap \cite{xia_2023_openearthmap}, a publicly available dataset designed for high-resolution land cover mapping. It includes $5000$ aerial and satellite images across 8 land cover classes, covering 97 regions in 44 countries across 6 continents. OpenEarthMap has been used in recent work such as \cite{xia_2023_openearthmap2}, which applies few-shot continual learning for land cover classification.

\section{Multi-modal data and challenge task} \label{sec:datatask}            
\subsection{The MapYourCity dataset} \label{sec:dataset1}       
\noindent In this work, the created dataset encompasses three modalities: Street-view images from Mapillary, Top-view satellite VHR Red-Green-Blue (RGB) images at 50 cm resolution, and Multi-spectral 12-band S-2 data at 10 meters.  
These input modalities have very different characteristics (i.e. number of spectral bands, spatial resolution, and building perspectives from top- and street-views). 
In the dataset, the \textit{three}     modalities are co-localized with respect to the specific building under study.  
In Figs. \ref{fig:short1} and \ref{fig:short2}, we show two examples with co-localized street-view and top-view satellite VHR modalities. In Fig. \ref{fig:short1vv54}, an example top-view VHR image and the corresponding S-2 image are shown, together with the building location in the centre, as a \textit{blue dot} in the image.                    



\begin{figure*}[t]               
  \centering         
  \includegraphics[width=7.9cm]{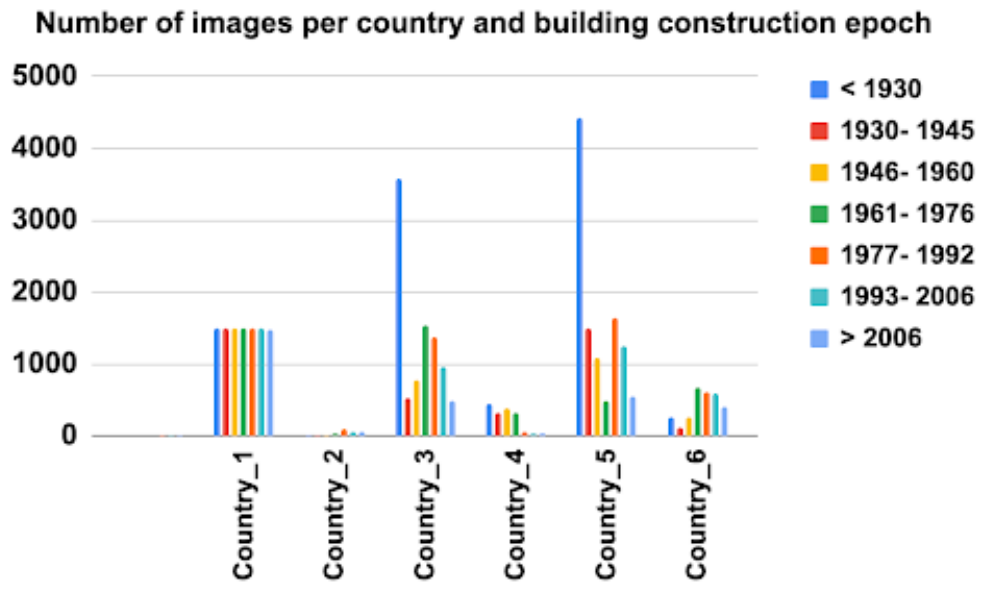}
  \includegraphics[trim={2.8cm 6.0cm 3.0cm 0},clip, width=9.3cm]{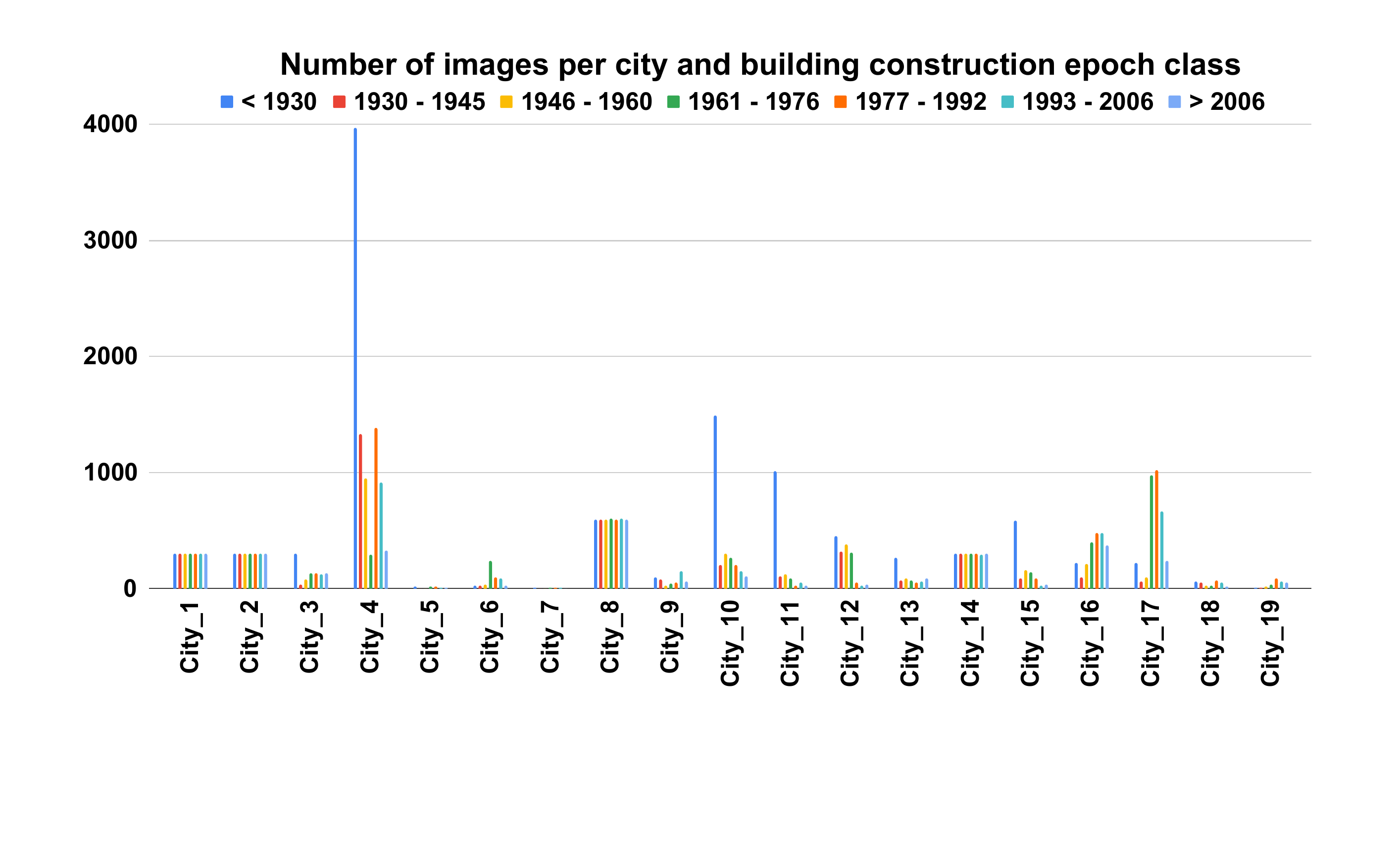}              
  \caption{Number of images per country and building age class (Left). Number of images per \textit{city} and building age class (Right).}         
  \label{fig:short4} \label{fig:short3}                   
\end{figure*}

The MapYourCity dataset includes cross-view imagery from $19$ cities across $6$ European countries, categorized into $7$ building construction epoch classes derived from the EUBUCCO database.        
The dataset contains images from different countries and cities, identified by the country ID and city ID.     
For model training, data from $15$ cities (with approximately $20000$ samples centered over specific buildings) were used, while the remaining $4$ cities (with approximately $5000$ samples centered over specific buildings) were held out exclusively for testing, enabling evaluation of generalization performance on previously unseen urban environments.      
These four cities are located in distinct countries (i.e. see Fig. \ref{fig:short4}), identified by the following country ID codes: QCD, HUN, FMW, and PNN.   
The building construction epochs in the MapYourCity dataset are categorized into seven classes, labeled from 0 to 6, based on the construction period with Class $0$: `Before 1930', Class $1$: `1930-1945', Class $2$: `1946-1960', Class $3$: `1961-1976', Class $4$: `1977-1992', Class $5$: `1993-2006', and Class $6$: `After 2006'.        
In Fig. \ref{fig:short3}, we show the number of image samples per country and building construction epoch class, as well as per city.  

The multi-modal MapYourCity dataset is released publicly with an open-source licence\footnote{\url{http://www.eotdl.com/datasets/AI4EO-MapYourCity}}.

\subsection{The MapYourCity data challenge} \label{sec:agebuil}                  
\noindent 
The primary objective of the challenge was to develop models capable of achieving high performance on previously unseen cities, i.e. on cities excluded from the training set. To evaluate generalization, the Private and Public Leaderboards include data from the above mentioned $4$ cities in a \textit{random} split, where these $4$ cities held-out for the test set have approximately $2100$, $100$, $1000$, and $1500$ samples, respectively. The public leaderboard reflected performance on a subset of the test data, while the private leaderboard used for final rankings was based on the remaining, unseen portion. The public leaderboard was continuously updated and visible throughout the four-month open competition, while the final private leaderboard was revealed only at the very end.\begin{figure}[]                                 
  \centering              
  \includegraphics[width=8.655cm]{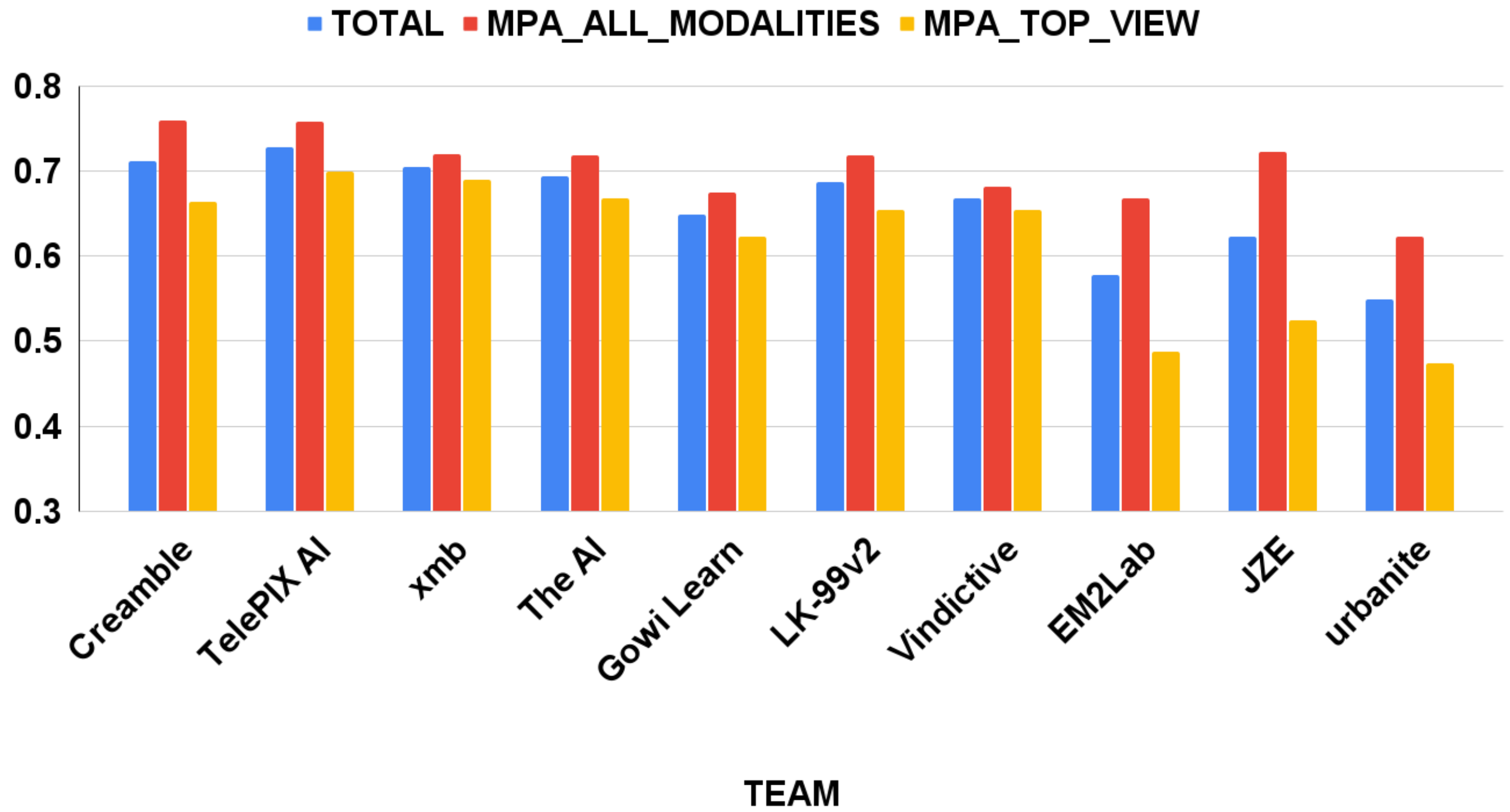}  
  \includegraphics[width=8.655cm]{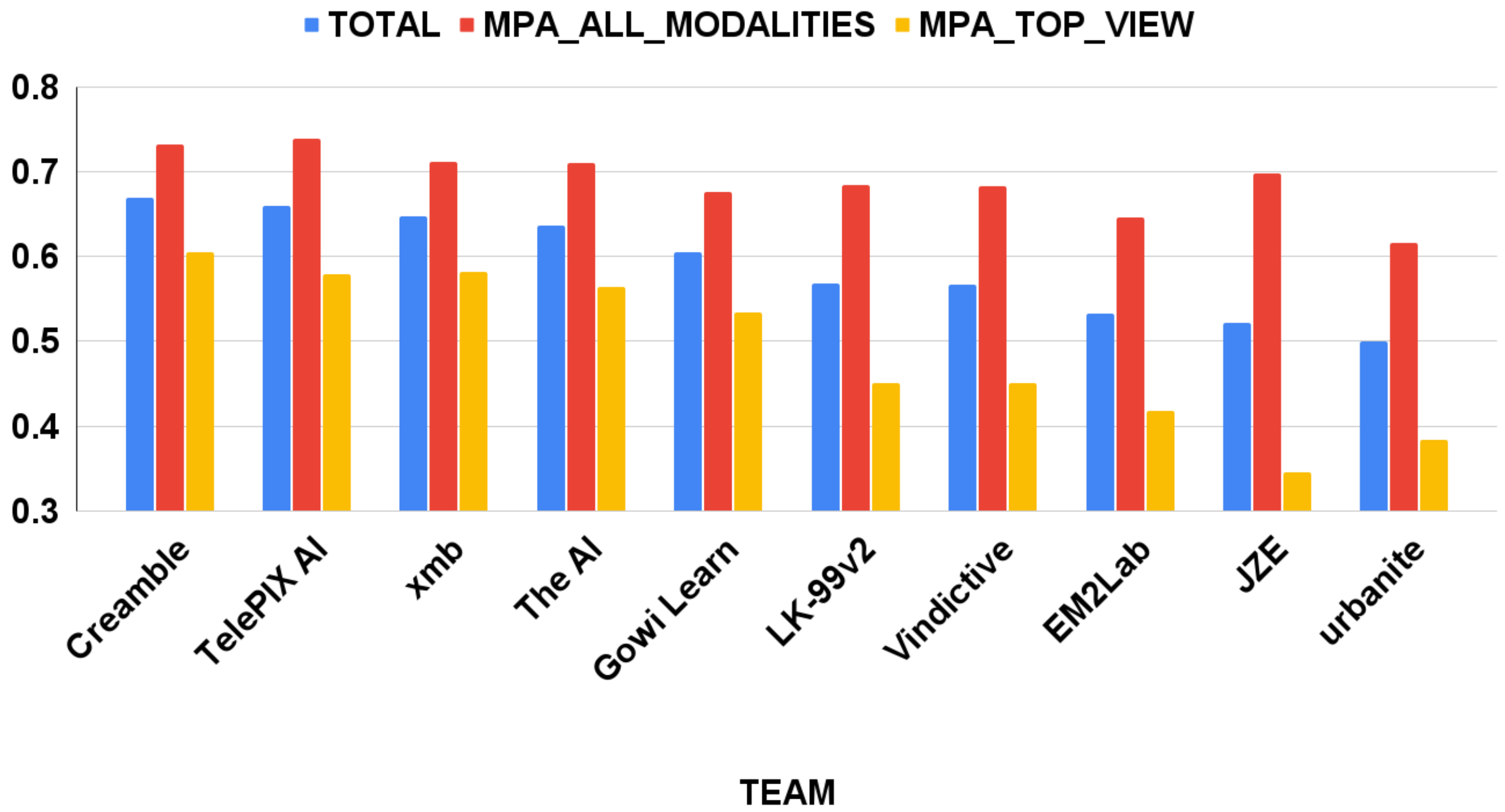}     
  \caption{Evaluation results of best models submitted via the open challenge, with public leaderboard (Top) and private final leaderboard of the challenge (Bottom).}    
  \label{fig:short5}     
\end{figure}

Additionally, half of the test samples excluded the street-view modality, enabling the evaluation of model robustness under missing modality conditions. We address scenarios where the street-view modality is missing during inference, a situation that frequently arises in practice. This consideration is crucial for ensuring the applicability of our approach in real-world deployment settings. Satellite imagery scales efficiently, enabling coverage of large geographic areas. In contrast, street-view images are less scalable, as capturing every building in a city manually is often impractical, time-consuming, and costly.

Through this challenge, we aim to advance research in EO Out-of-Distribution (OoD) generalization and multi-modal learning with incomplete data, two critical areas for real-world deployment of AI models in EO and urban analytics. Hence, we present evaluation results for the EO OoD generalization performance \cite{stratificationpaper} of the top-performing models from the challenge under \textit{two} distinct settings:  
\begin{itemize}
    \item Full modality setting: Both training and inference are performed using all three modalities of the MapYourCity dataset.  
    \item Top-View only setting: Training is conducted using all three modalities, but inference is performed using only the top-view satellite modalities (VHR and S-2), excluding the street-view images.
\end{itemize}


The challenge was open to the entire community for four months in 2024\footnote{https://platform.ai4eo.eu/map-your-city}. A total of $123$ teams registered for the competition. Among them, $30$ teams actively submitted solutions and were ranked on the leaderboard. A total prize pool of €$5,000$ was awarded to the top three entries of the MapYourCity challenge\footnote{Videos are in: \url{http://www.youtube.com/watch?v=oVfINRBFIUY} and \url{http://www.youtube.com/watch?v=z5jCkmP8Az8}. In addition, the webinar video recording is in: \url{http://www.youtube.com/watch?v=P_fr4IwC4iA}.}.    
To assess the performance of each submitted model, we calculate the average of the \textit{diagonal} items of the confusion matrix, known as the Mean Percentage Accuracy (MPA). We present the evaluation results of the models from the MapYourCity challenge in Fig. \ref{fig:short5}. The full leaderboards are also available publicly\footnote{\url{https://platform.ai4eo.eu/map-your-city/leaderboard}}.

\begin{figure*}[t]                              
  \centering              
  \includegraphics[width=14.2cm]{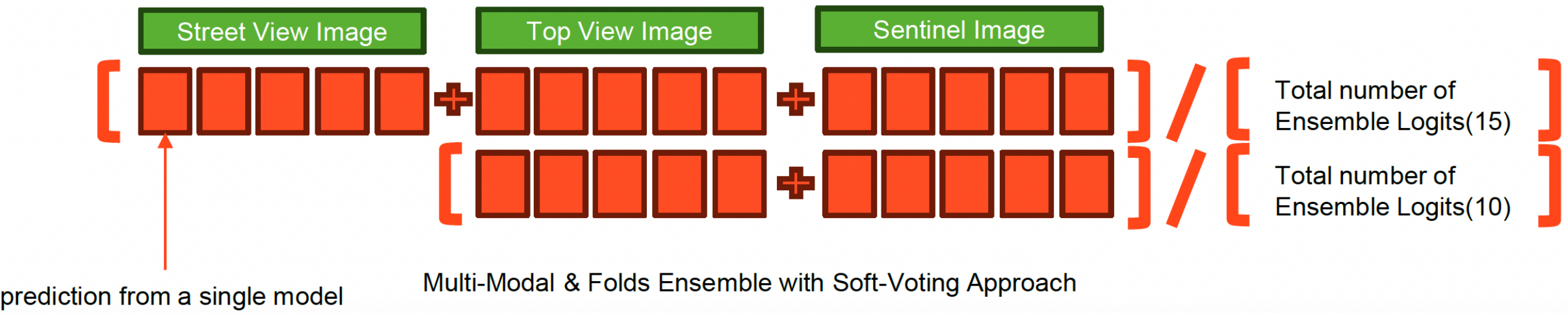}           
  \caption{Multi-Modal and Folds Ensemble with Soft-Voting Approach: model proposed by Team Creamble.}      
  \label{fig:short1aaaasfgasg1}          
\end{figure*}


Overall, the highest score achieved in the Full modality setting on the private leaderboard is $0.74$, demonstrating that estimating the construction epoch of buildings in unseen cities is feasible using the proposed models. Performance in the Top-View setting is notably lower compared to the Full modality. When relying solely on top-view data during inference, model accuracy drops by more than 10 points in the MPA score, highlighting the importance of multi-modal inputs for robust estimation. However, using only EO data, without street-level imagery, still enables reasonably good performance, underscoring the value of EO data as a scalable and globally available resource. It is also worth noting that performance on the private leaderboard is significantly lower than on the public leaderboard, indicating a certain lack of generalization capabilities, particularly in the Top-View only setting\footnote{The methods and models are also presented in: \url{http://urbis24.esa.int/wp-content/uploads/2024/10/01_AI4EO_MapYourCity_URBIS.pdf}. In addition, the baseline model is also presented in: \url{http://drive.google.com/file/d/1dcNLswBlC-bS85H5mi3K4CFou9J6FFfZ/view?usp=sharing}.}.         


In the following sections, we present and analyze the best models submitted to the challenge. Specifically, we focus on the Top-4 performing models, which achieved the highest accuracy on the held-out test cities. To achieve \textit{top} performance, the models employ different deep learning techniques, for example, 
class label correlation modelling,  
choosing the best-performing model architectures, 
using a different encoder for feature extraction for each modality 
or         
a \textit{shared} encoder to align and share features from different modalities, 
or training different models for different countries.  
Moreover, the top models perform ensemble techniques, 
weight the importance of features,   
employ K-fold Cross Validation,     
perform \textit{late}     fusion to combine the extracted features, 
combine different feature fusion methods to learn feature importance, i.e. feature concatenation and/or geometric data fusion, 
ensemble the outputs of different trained models across different folds, 
or    
employ a hide-out strategy for the street-view images to ensure model robustness.  

\begin{figure} 
    \centering       
    \includegraphics[width=1.0\linewidth]{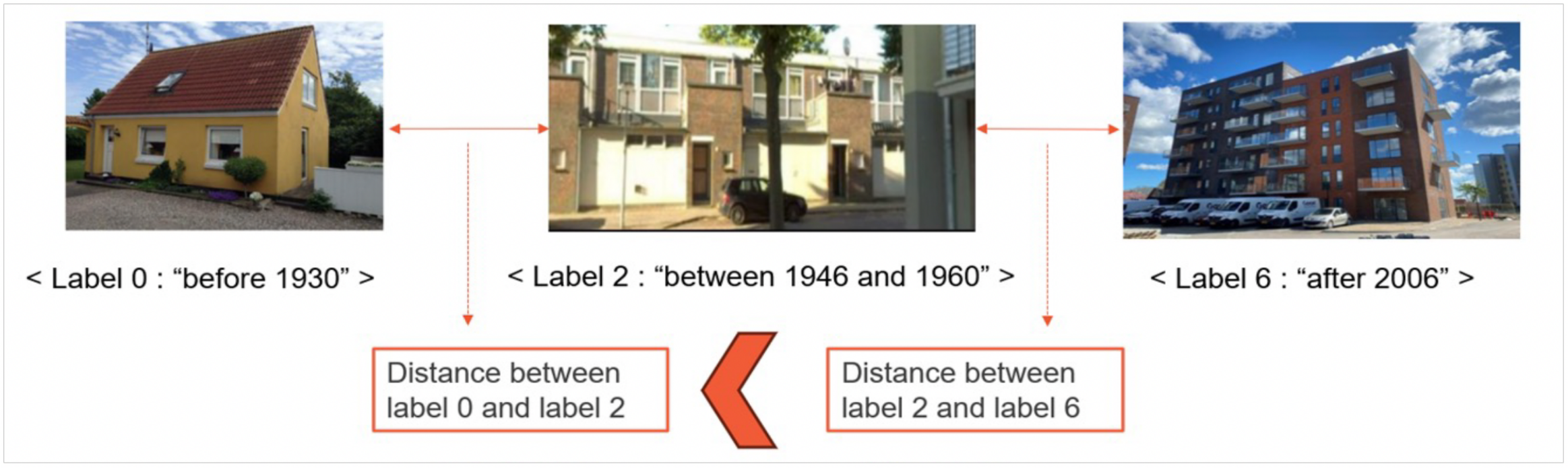}  
    \caption{Temporal proximity: Correlation between labels.}\label{fig:fignewfigdistlabels}   
\end{figure}


These models are examined in terms of their data preprocessing strategies, architectural design, ensemble approach, hyperparameter optimization, experiments and performance results under different evaluation settings. 
In addition to the top-performing submissions, we also present the baseline model of the challenge. This model, based on recent Transformer architectures \cite{transf3}, \cite{vitdosovitskiy}, was provided to all participants as part of the starter toolkit to serve as a reference point for model development and performance comparison.

\section{$1^{st}$ Ranked Solution: Team Creamble}\label{sec:secfito}\label{sec:mainmodels}       

\noindent This section presents the top-ranked solution to the MapYourCity Challenge. The proposed model is outlined in Fig. \ref{fig:short1aaaasfgasg1}.


\subsection{Data Preparation}\label{sec:labelcorrsec}          

\noindent The task of the MapYourCity challenge inherently involves a strong correlation between class labels because of their temporal proximity. 
For example, Class 6 (`after 2006') is more closely related to Class 5 (`1993–2006') than to Class 0 (`before 1930'). Therefore, recognizing and incorporating these temporal relationships and correlations is essential for the model to learn effectively and make accurate predictions about the age of the buildings (see Fig. \ref{fig:fignewfigdistlabels}).   

\begin{figure*}      
    \centering          
    \includegraphics[width=0.85\linewidth]{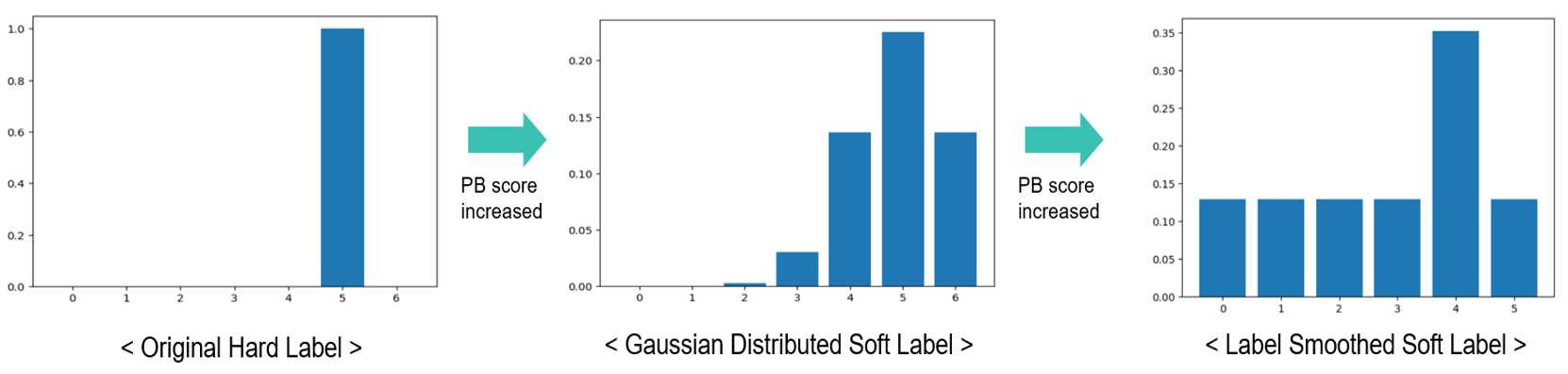}  
    \caption{Label correlation modelling in the MapYourCity dataset, by Team Creamble.}\label{fig:figureprobablabel}    
\end{figure*}

   
Our first approach is to assume that the label distribution is probabilistic, rather than binary.    
Soft labels (rather than hard labels) are used for the supervised learning classification training of the model \cite{softlabelspaper, vriespaper}.     
We convert the labels into soft labels by using a Gaussian probabilistic distribution, with the learning guided by the KL-Divergence Loss, in Fig.~\ref{fig:figureprobablabel}. This approach yields a slight improvement in performance. However, manually tuning the variance parameter introduces uncertainty and additional cost.   
To address this, we implemented Label Smoothing \cite{whendoeslabelsmooth}, achieving our best results with a smoothing factor of $0.3$, identified through hyperparameter optimization using Optuna.           

As illustrated in Figs.~\ref{fig:figureprobablabel} and \ref{fig:fignewfigdistlabels}, the Gaussian-based approach to modeling label correlation is both intuitive and effective. It captures the temporal proximity between classes, enhancing the model's ability to generalize across adjacent time periods. Incorporating the correlation between class labels, based on their temporal distances, into the learning process is crucial for achieving high performance. Our results and numerical analysis demonstrate that the combination of label smoothing and Gaussian-based label correlation significantly enhances classification accuracy. Notably, this improvement is not observed when either technique is applied in isolation, highlighting the importance of their joint application, i.e. significant improvement in performance\footnote{Label correlation modelling: \url{https://github.com/jeantirole/MultiModal_Lab/blob/master/00.Fundamentals/26.Label_to_Distribution_v3.ipynb}.}.

To assess our label‐correlation strategy, we evaluated three approaches on the Top-View validation split: 1) standard cross‐entropy loss training (recall $0.69$), 2) incorporating inter‐label correlations via a Gaussian prior (recall $0.71$), and 3) uniform label smoothing (recall $0.72$).        
Relative to the baseline, Gaussian correlation modeling yielded a $2.01$-point increase in recall, while label smoothing delivered a $3.01$-point gain.     
These results demonstrate that embedding prior knowledge, specifically label co-occurrence and spreading the probability mass across all classes, boosts generalization and serves to penalize overconfident outputs, and as we will also see in Table \ref{tab:maintablesumm} (i.e. see Sec. \ref{sec:discdiscussion}), the label‐correlation strategy is a main crucial feature of the model.

\begin{figure} 
    \centering     
    \includegraphics[width=1.0\linewidth]{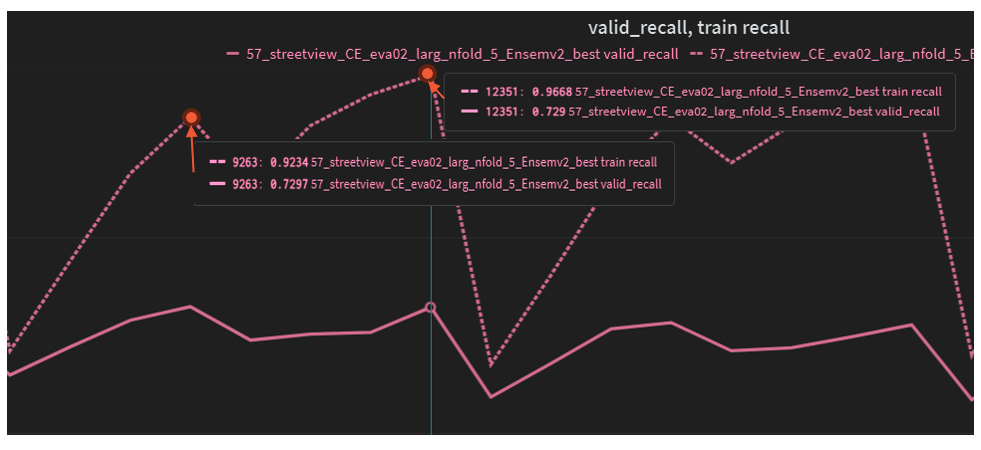} 
    \caption{Street-view best model training logs for Team Creamble.}\label{fig:figlogs}  
\end{figure}

To resize the various street-view images, we first analyze the distribution of their original dimensions. Based on this analysis, we select a $2$:$3$ aspect ratio, which best represents the overall distribution and minimizes information loss during resizing. For top-view images, which naturally have a $1$:$1$ aspect ratio, we follow the training size of the model during resizing.

\subsection{Models}          

\noindent 
We evaluate several high-performing models on ImageNet for image classification tasks. Specifically, we focus on Transformer-based architectures such as EVA \cite{EVANEW, EVA2}, BEiT \cite{beit}, and MaxViT \cite{maxvit}. While these models demonstrate strong overall performance, their effectiveness varies across different image modalities (i.e. street-view, top-view VHR, and S-2 imagery). This variation is likely influenced by differences in the datasets and the pretraining strategies employed (e.g., supervised and/or \textit{self}-supervised). Ultimately, we identify EVA-02 \cite{EVA2} as the most suitable model for this task, as it delivers superior feature extraction performance on the multi-modal MapYourCity dataset, considering both its parameter size and the nature of the input images. 

We apply a soft-voting ensemble approach to accommodate the test dataset's structure with the Full modality and the Top-View only settings. Additionally, ensemble layers are trained by \textit{freezing} domain-specific models and focusing on ensembling the logits, which we find that significantly improves the model's final decision-making process.       
The method, depicted in Fig. \ref{fig:short1aaaasfgasg1}, efficiently handles \textit{both} cases and incorporates K-fold cross validation.    
This avoids the need for additional layers to combine multi-modal data, improving efficiency and reducing overfitting on the public leaderboard.


\begin{figure*}
  \centering        
  \includegraphics[width=15.51cm]{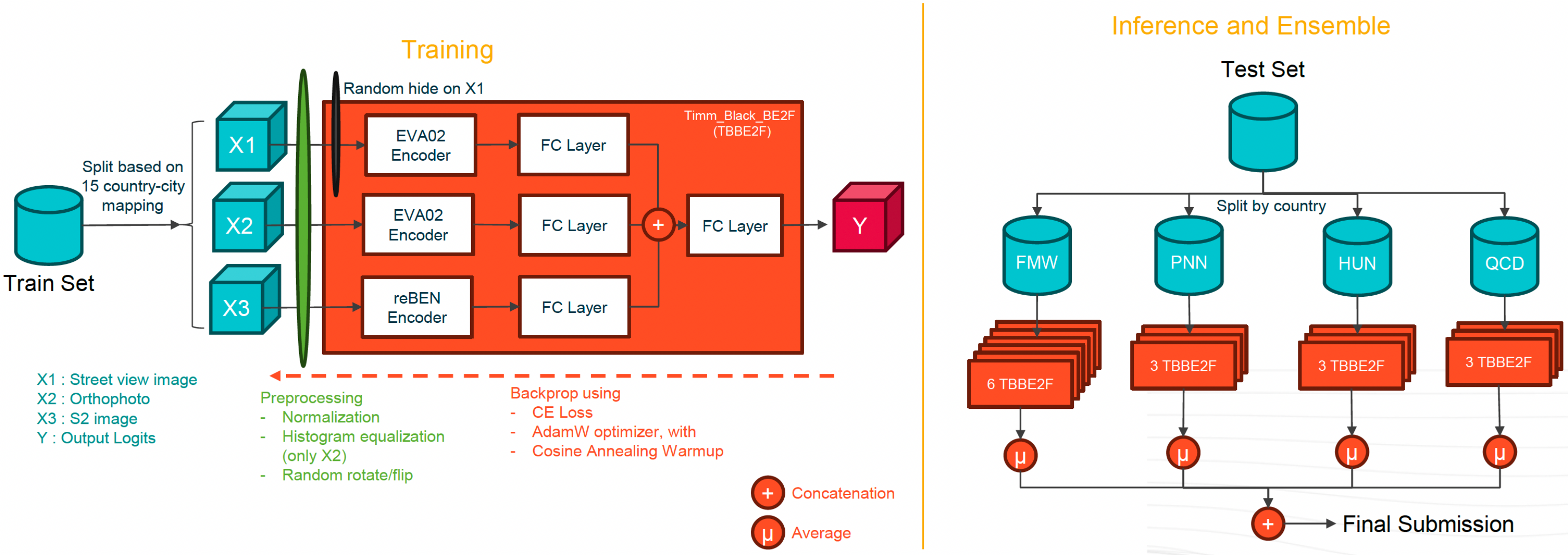}      
  \caption{Overall diagram of the approach proposed by Team TelePIX AI.}    
  \label{fig:short1aaaasfgasg2}          
    \label{fig:overall-diagram} 
\end{figure*}\begin{figure} 
    \centering    
    \includegraphics[width=\linewidth]{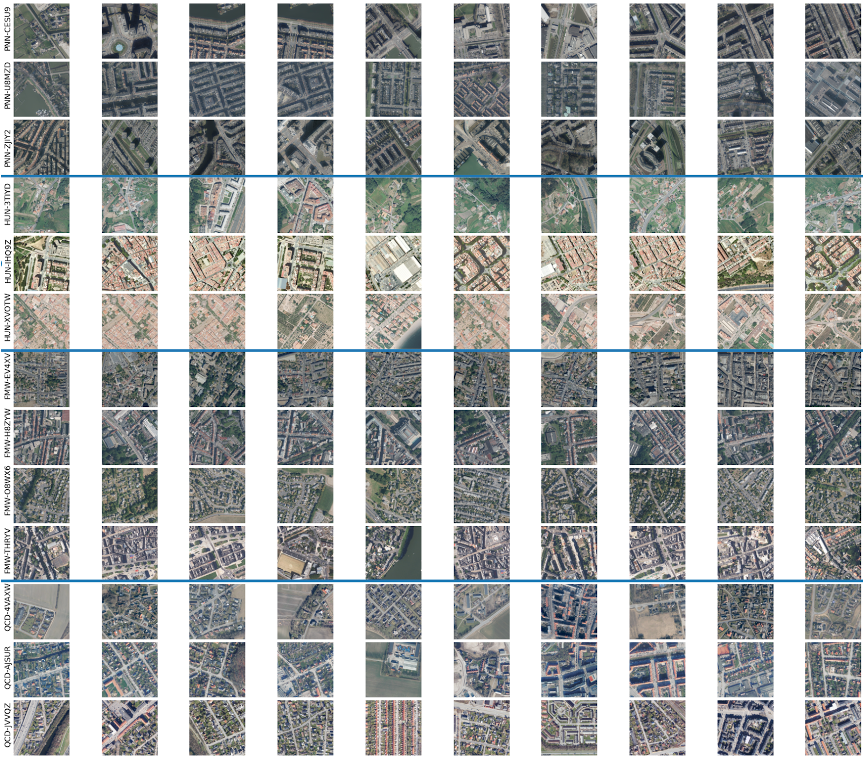} 
    \caption{Ten sample top-view VHR images from each city in the same country (separated by a blue line), where from the top, we have: PNN, HUN, FMW, and QCD.} 
    \label{fig:sample}           
\end{figure}

\subsection{Results and Discussion}       

\noindent We utilize K-fold Cross Validation and integrate it into the ensemble algorithm. The competition’s MPA final evaluation metric, which often closely resembles recall, is used as the criterion to select the \textit{best} model.

During optimization, we observe that the highest validation recall values appear at two points during training (see Fig. \ref{fig:figlogs}): i) early in epoch $4$, and ii) \textit{late} in epoch $7$.       
The difference between these \textit{two} points is    due to the \textit{gap} between training and validation recall.    
For example, the validation scores at epochs $4$ and $7$ were $0.7297$ and $0.7290$, respectively, while the training recall values were $0.92$ and $0.96$.  
This suggested potential overfitting in \textit{later} epochs.   
However, this pattern was inconsistent across folds. 
To mitigate this, we employ CosineAnnealingWarmRestarts, allowing us to adjust the learning rate regularly, preventing local minima and overfitting. \textit{AdamW} was also used in the model training to improve generalization performance.

Among the ensemble methods, the combination of the results of regression and classification using Cross-Entropy is also successful. The Euclidean distance from the MSE model predictions is converted into a percentage and incorporated into the Cross-Entropy Softmax values, \textit{boosting} performance.   
However, this ensemble structure is    unfortunately not included in the final submission. Nonetheless, there is potential for its application in problems that can be interpreted and tackled from \textit{both} regression and classification perspectives, where labels have some degree of correlation.

{\color{black}The Team Creamble model therefore builds upon EVA-02 \cite{EVA2}, where we note that the number of parameters is approximately $300$M. In addition, the FLOating Point operations (FLOPs) are approximately $360$G FLOPs, and the number of ensembles used is $15$, i.e. $5$-fold times $3$ variations for street, high-resolution remote sensing, low-resolution remote sensing. 
We also examined the correlation between the model performance and the computational complexity of the models.    
The performance among the top ViT models did not scale linearly with model size or complexity. More specifically, EVA-Large (approximately $300$M parameters) achieved comparable or better results than larger models like EVA-Giant (approximately $1$B parameters), indicating diminishing returns with increased complexity. We also note that models based on ResNet CNN performed worse compared to large models built on ViT. The Team Creamble model during training uses image size $448$ $\times$ $448$, i.e. relatively large images, batch size $8$, and uses two NVIDIA A6000 GPU machines where the training time was $6$ hours.}

The code for the solution is available openly\footnote{\url{https://github.com/jeantirole/Challenge_MapYourCity}}.


\section{$2^{st}$ Ranked Solution: Team TelePIX AI}\label{sec:sectop}     

\noindent This section presents the second-top solution to the MapYourCity Challenge\footnote{As the participants were affiliated with the same organization as Team Creamble, the Team was not eligible for a cash prize in order to maintain fairness in the competition.}. The overall architecture and workflow are illustrated in Fig.~\ref{fig:short1aaaasfgasg2}. 

\subsection{Data Preparation}\label{sec:newusenewdataprep}         
\noindent In the proposed approach, data preparation aims to reduce the complexity of the task of the challenge. The data are \textit{split}    for training and validation using K-Fold cross-validation, with \textit{each fold representing a city within the respective country}. For the countries in the test set —FMW, PNN, HUN, and QCD— $6$, $3$, $3$, and $3$ folds are created, respectively, with the number of folds varying based on the number of cities in each country (FMW had the most).   
For each fold, $2$-$3$ cities are used for training, while $1$ city is reserved for \textit{validation}.    
Several preprocessing techniques are applied to the input images. These include normalization, performed by subtracting the mean and dividing by the standard deviation, for all three modalities. Additionally, histogram equalization is applied to the to-view VHR images to enhance contrast, and data augmentations such as random rotation and flipping are also used to improve model generalization.

\subsection{Models}      
\noindent 
VHR images from cities within the same country (i.e. Fig.~\ref{fig:short3}) tend to appear \textit{similar}, as it can be seen in Fig.~\ref{fig:sample}, while images from different countries show significant variation. To take advantage of this important characteristic, the model is \textit{trained separately for each country} using City-Fold cross-validation for data splitting to minimize the complexity of the patterns (i.e. discriminative features) that are needed to be learned. The model, referred to as TBBE2F, processes each input image through \textit{pre-trained encoders that are fine-tuned during training}.  

The model processes three input images: street-view, top-view VHR, and S-2. Each input is handled by its own dedicated pre-trained encoder. 
Given the distinct domains of these images, separate encoders are assigned to each modality.  
During training, these \textit{encoders are fine-tuned to enhance feature extraction}. In addition, for classification, the cross-entropy loss is employed.  

The extracted features from each modality are represented as vectors, with their dimensions depending on the encoder architecture used.
\textit{A variety of encoders are explored}, including SeResNext \cite{hu2019squeezeandexcitation}, 
MobileNet \cite{howard2017mobilenets}, 
EfficientNet \cite{tan2019efficientnet}, 
and several ViT-based \cite{dosovitskiy2020vit} encoders pre-trained variations, 
along with EVA02 \cite{EVA2} and reBEN \cite{clasen2024reben}, which are combined across the three inputs. 

After feature extraction, the resulting vectors are passed through a Fully-Connected (FC) layer, producing seven logits (i.e. corresponding to the number of classes). Then, these logits are concatenated and passed through another FC layer, reducing the $21$ logits back to seven. Hence, the team TelePIX AI employs a prediction-level data fusion strategy, as opposed to for example fusing data at the feature level.   
The final predictions are processed through a \textit{SoftMax function to obtain the class output probabilities}.      

In addition, a \textit{hide-out strategy} is applied during training for the street-view input. The latter is important and is specifically designed to address the missing modality problem during inference, where street-view images are unavailable. With a probability of $0.5$, the stret-view image is blacked out (i.e. multiplied by $0$), encouraging the model to avoid over-reliance on this modality and maintain the performance when it is absent.  
This approach also allows the model to focus on learning important features from top-view modalities, while using the street-view pictures as a guide.

The training process results in $15$ models, each corresponding to a fold from one of the four countries as described in Sec.~\ref{sec:newusenewdataprep}. Inference is conducted using these models on the test datasets relevant to the country in which each model was trained on. The outputs for each country are averaged, as depicted in Fig. \ref{fig:overall-diagram}.

\begin{figure}[t]        
    \centering         
 \includegraphics[width=9.525cm]
 {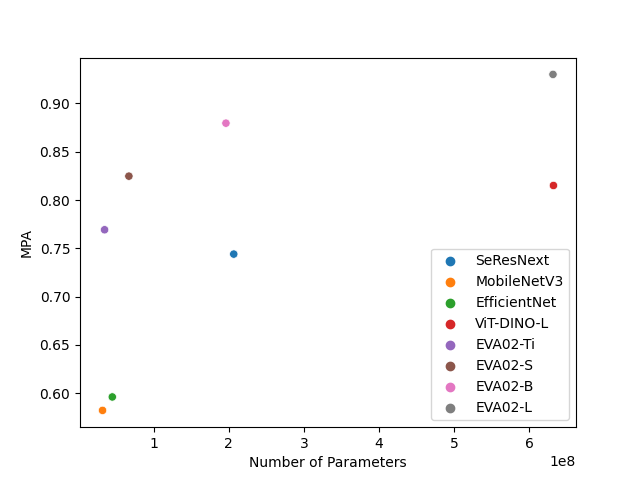}   
    \caption{Performance analysis for street-view and VHR encoders for model selection, by Team Telepix AI.}   
    \label{fig:encoder}   
\end{figure}\begin{figure*}[t]                     
  \centering           
  \includegraphics[width=14cm]{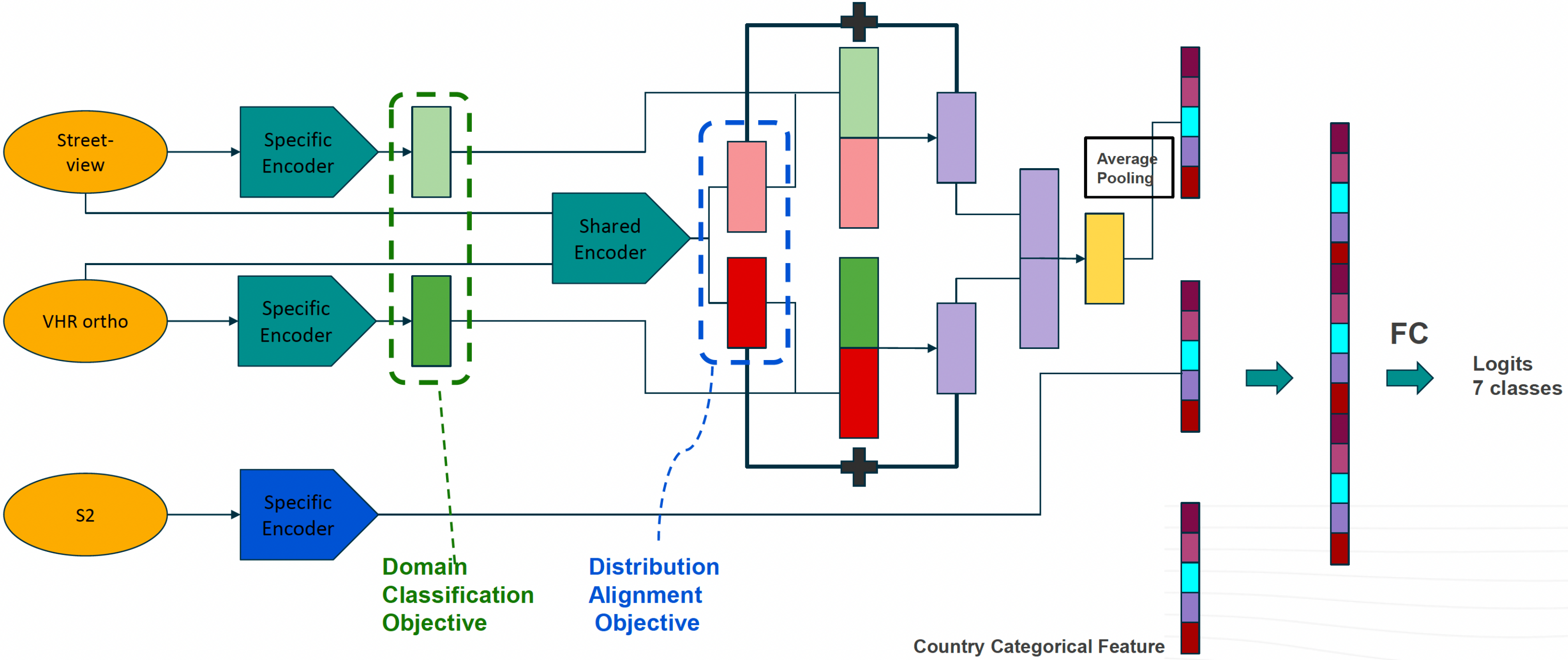}      
  \caption{Flowchart diagram of model using an additional shared feature encoder for the street- and top-view VHR images and the distribution alignment loss, by  Team xmb.}         
  \label{fig:short1aaaasfgasgasdf1}             
\end{figure*}

\subsection{Results and Discussion} 

\noindent To identify the optimal solution, the first step involves selecting the most suitable encoder for each input modality. This is achieved through a controlled evaluation process: the encoder for the street-view input is varied while keeping the encoders for the top-view fixed; then, the encoder for VHR images is varied with the others held constant; and finally, the encoder for S-2 is varied while maintaining the encoders for the others. This systematic approach allows for isolating the impact of each encoder on overall model performance.   
Once the best combination of encoders is identified, \textit{Optuna is employed to perform hyperparameter search}. The optimal hyperparameters, identified through the Optuna search, are     as follows: a learning rate of $1.32$, a batch size of $8$, warm-up steps of $300$, and a decay factor of $1.1$.     
The optimal combination is then used for the inference in the test dataset, and the results are ensembled, as outlined   in Fig.~\ref{fig:short1aaaasfgasg2}, to generate the final submission.


The experimental results show that the encoder that performs well for street-view images also performs well for VHR images. A comparison of the performance of each encoder type can be seen in Fig. \ref{fig:encoder}. The EVA02 encoder emerges as the \textit{best}, with a significant margin over the other architectures. Given that larger models usually tend to yield better performance, we select the \textit{EVA02-L} variant for both street-view and VHR modalities. For S-2 input, due to its distinct number of channels (12 multi-spectral bands), the best-performing encoder is \textit{reBEN}, which is capable of leveraging all available channels effectively. 

\textit{The experiments also reveal that} random hiding of street-view images improve the overall MPA     by $2$-$3$\%, ensembling boosted it by $1$-$3$\%, and pre-processing with histogram equalization contribute an additional $1$-$2$\% improvement.  
As we will also see in Table \ref{tab:maintablesumm} (i.e. see Sec. \ref{sec:discdiscussion}), the random hiding of street-view images strategy is a main important feature of the model.    
Ultimately, the proposed solution achieves MPA $73.89$\% on the private leaderboard for all modalities, and MPA $57.97$\% for the Top-View only setting. These results demonstrate the effectiveness of the proposed encoder selection, hyperparameter tuning, and ensemble strategy.


Key factors contributing to the model’s strong performance include: i) robust encoders (e.g., EVA-02 and reBEN), ii) random hiding of street-view images with a probability of $0.5$, encouraging the model to learn discriminative features from VHR and S-2 imageries, iii) country-based data splitting to ensure generalization across geographic regions, and iv) ensemble techniques and comprehensive pre-processing, which improves both training stability and final prediction robustness.

The code for the solution is available openly\footnote{\url{http://drive.google.com/drive/folders/1fuc4UYbhQBtHeN2rtHKMBh9BxCRRFVAi?usp=sharing}}. 

\section{$3^{rd}$ Ranked Solution: Team xmb}\label{sec:sebe}   

\noindent This section presents the third-best solution to the MapYourCity Challenge.  

The proposed solution is outlined in Fig. \ref{fig:short1aaaasfgasgasdf1}.  

\subsection{Data Preparation}      

\begin{figure}
    \centering               
    \includegraphics[width=3.8in, height=1.5in]{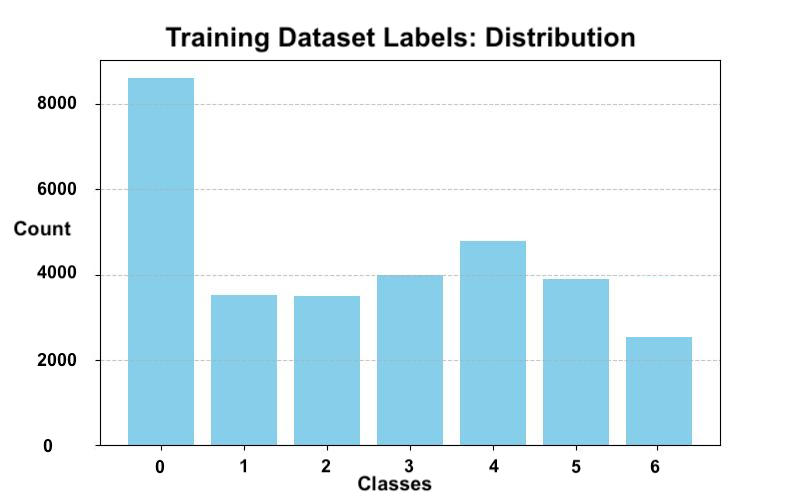}   
    \caption{Distribution of the training dataset labels.}   
    \label{fig:labels_dist}        
\end{figure}

\noindent As shown in Fig.~\ref{fig:labels_dist}, the training dataset exhibits a slight class imbalance, with label $0$ appearing approximately twice as often as the other labels. 
The MapYourCity multi-modal dataset includes seven classes, each representing a range of building construction years. The dataset exhibits class imbalance, a common and realistic scenario in real-world applications. In practice, it is typical to encounter uneven distributions across categories, especially in historical data, as buildings are constructed across different years, time periods, and architectural eras.
To mitigate this class imbalance, we employ the Focal Loss during training across all models \cite{focallosspapercite, focalloss2B, focalloss3C}. The distribution of samples per category directly impacts model accuracy and robustness. In particular, the imbalance in this multi-modal dataset can lead to overfitting on the underrepresented classes. By using the Focal Loss, we reduce this effect and improve generalization, especially for the classes with limited data. 

Both the training and test sets contain information from six different countries. Since architectural styles can vary significantly between countries, potentially offering clues about the age of buildings (e.g., Fig. \ref{fig:sample}), we convert the country index into a one-hot encoded vector and use it as a categorical feature. We note that we deliberately avoid applying this approach to cities, as the test dataset may include held-out cities that were not seen during training.

\subsection{Models}        
\noindent We     propose here \textit{two}     types of classification models. 

\begin{figure*} 
    \centering    
    \includegraphics[height=1.45in]{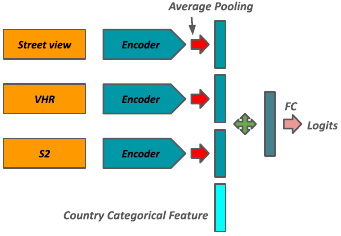} 
    \includegraphics[width=4.8in]{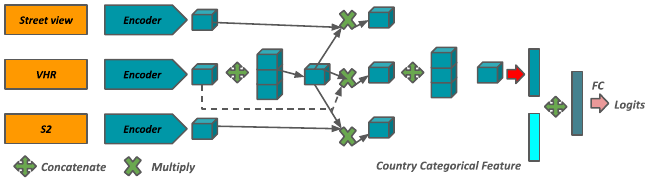}    
    \caption{Feature Concatenation (Left) and Geometric Fusion (Right) modules, proposed by Team xmb.}   
    \label{fig:geometric_fusion}\label{fig:feature_concatenation}          
\end{figure*}

For the Full modality setting, both training and inference are carried out using all available modalities. The inputs are processed through three encoders of the same feature extraction model, and then combined by late fusion. Additionally, we also employ \textit{two}    data fusion methods: Feature Concatenation, and i) Geometric Fusion (see Fig. \ref{fig:geometric_fusion}). In \textit{Feature Concatenation}, the average pooling features are combined with the country categorical feature and, then, passed into the final fully-connected layer. Inspired by \cite{chen2022multi}, in \textit{Geometric Fusion}, features before average pooling are concatenated and processed through a CNN to learn feature importance. Next, the resulting weights are then multiplied by the inputs and also concatenated again to produce the final output features.

\begin{figure*} 
    \centering       
    \includegraphics[height=1.80in]{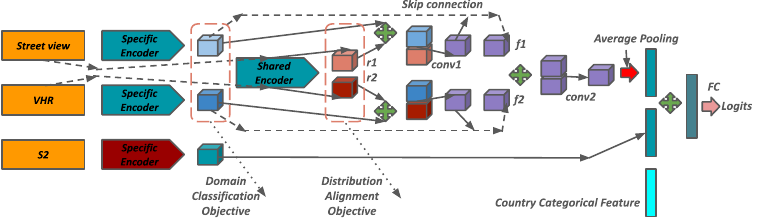}   
    \caption{Shared-Specific Feature Modelling for the Top-View only setting, developed by Team xmb.}      
    \label{fig:missing_modality}        
\end{figure*}

For the Top-View only setting, training utilizes all modalities, while inference is based solely on the two top-view inputs. Inspired by \cite{wang2023multi}, we adopt a \textit{Shared-Specific Feature Modelling approach} to address the missing modality issue (see Fig. \ref{fig:missing_modality}).     
Each modality is assigned a \textit{specific encoder}, with street-view and VHR inputs that also share an encoder.   
Here, the \textit{shared encoder} aligns the features of the VHR and street-view.     

During inference, aligned VHR features are used in place of street-view features. The distribution alignment loss function term \cite{MissingModality1} is also added to the objective function that is minimized during training. 
The street-view and VHR modalities use the same encoder for both shared and specific encoders, while the S-2 modality uses a different, shallower encoder. We also apply input dropout, randomly replacing street-view images with all-zero images (e.g., like the model described in Sec.~\ref{sec:sectop}).

\subsection{Results and Discussion}                 

\noindent We implement     our approaches using the PyTorch library, the PyTorch Lightning framework, and timm   \cite{rw2019timm} library for the feature encoders. The efficient and lightweight encoders from timm were chosen for our experiments, with Efficientnetv2\_s,  Mobilevitv2\_150, and Efficientnetv2\_b3 for the Full modality setting, and Efficientnetv2\_b1 (for street-view and VHR) and Efficientnetv2\_b0 (for S-2) for the Top-View only setting.

We train $4$ models for the Full modality setting and use their ensemble as the final model. 
Also, only $2$ models are trained for the Top-View only setting.  For most of the models, we use a batch size of $40$.    
The models are trained using the AdamW optimizer (i.e. learning rate: $1e-4$, weight decay: $1e-3$) and a CosineAnnealingLR scheduler (i.e. $\eta_{min}$: $5e-5$) for $30$ epochs without early stopping, also applying the Focal Loss with Label Smoothing (i.e. with a smoothing coefficient of $0.1$) \cite{focallosspapercite}, \cite{whendoeslabelsmooth}.        

The train dataset is split into $10$ stratified folds, with each model using one fold for validation and the remaining $9$ for training. In addition, the street-view and VHR inputs are resized to $512$x$512$ and normalized to [$0$, $1$].       
Moreover, the S2 inputs are clipped at a maximum value of $10,000$ and normalized to [$0$, $1$].      
Using the Albumentations library, we apply data augmentations such as CoarseDropout, GridDropout, and \textit{Spatter} to    enhance the robustness of the street-view input during training. We also apply test-time data augmentation for both model types using Flip transformations.     

Our model submission achieves a MPA score of $71.22$ on the private leaderboard when all modalities are used. When limited to top-view modalities only, the MPA score drops to $58.14$. These results demonstrate the effectiveness of the shared-specific feature modeling approach (i.e. Figs.~\ref{fig:short1aaaasfgasgasdf1} and \ref{fig:missing_modality}) in addressing the missing modality issue.



In the full modality setting, only the features at the end of the encoders are fused by late fusion, leaving the intermediate-layer features unconnected. Exploring middle fusion, where features from intermediate layers are integrated, could be a valuable direction for future work. Additionally, incorporating an attention mechanism for data fusion may further enhance performance by enabling more dynamic and context-aware integration of modalities.

The code for the solution is available openly\footnote{\url{http://github.com/xmba15/ai4eo_map_your_city}}.

\section{$4^{th}$ Ranked Solution: Team The AI Buzzard}\label{sec:thibes}      

\begin{figure*}[t]                       
  \centering        
  \includegraphics[width=15.6cm]{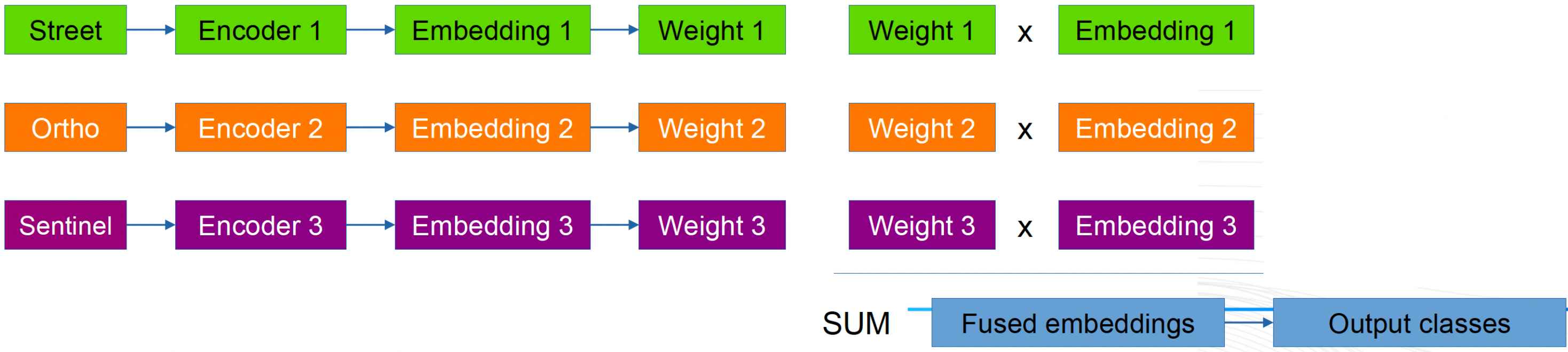}     
  \caption{Flowchart diagram of the model that uses SwinT encoders, developed by team The AI Buzzard.}      
  \label{fig:short1aaaasfgasgasdf2}           
\end{figure*}

\noindent This section presents the fourth-best solution to the AI4EO MapYourCity ESA Challenge. The flowchart diagram is shown in Fig. \ref{fig:short1aaaasfgasgasdf2}.    

\subsection{Data Preparation}            

\noindent  
We split the development set with $1000$ samples from the training dataset, following the distribution of country IDs in the test dataset. We apply data augmentation. All VHR images are resized to fit the expected resolution of the model. We apply random horizontal and vertical flips with a probability of $p=0.5$, as well as color jitter, \textit{contrast}, varying brightness, and saturation.   
All street-view facade photos are randomly cropped and resized during training, and also resized during inference.    

S-2 multi-spectral images are usually processed to fit standard computer vision models. Here, the original images (i.e. $64\times 64$ pixels, $12$ channels) from the MapYourCity multi-modal dataset are rearranged to size $128\times 128$ pixels and $3$ channels (RGB): in addition, the top left corner contains the NDVI, NDWI, and NDBI values; the top right the B04, B03, and B02 bands; the bottom left the B05, B06, and B07 spectral bands; and the bottom right the B8A, B11, and B12 bands, respectively.

\begin{figure} 
    \centering            
    \includegraphics[width=0.92\linewidth]{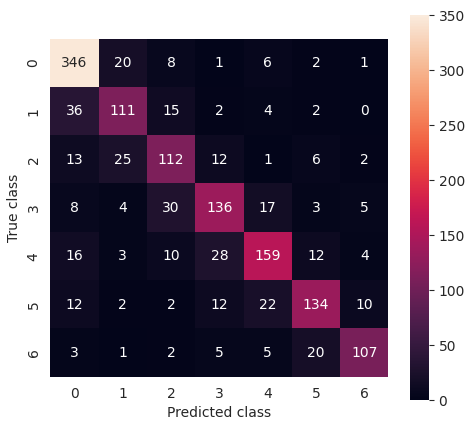}
    \caption{Confusion matrix for the Full modality setting, from the model by team The AI Buzzard.}
    \label{fig:confusionmatrix}             
\end{figure}

\subsection{Model}              

\noindent We     develop a flexible and robust multi-modal model. Each input modality (street-view, VHR, and S-2) is processed by a \textit{pretrained SwinV2 encoder} \cite{liu2022swintransformerv2scaling}, \cite{vitdosovitskiy} from the timm collection \cite{rw2019timm}. These models are pretrained on the ImageNet dataset     to extract general image features.      
Here, at this point, we note that as we will also see in Table \ref{tab:maintablesumm} (see Sec. \ref{sec:discdiscussion}), the use of the recent SwinV2 Transformer model is a main significant feature of the model.   

The resulting embedding vectors are combined by late fusion \cite{multimodalgit}. Each embedding is processed through a linear layer that yields a learning attention weight. Then, as also illustrated in Fig. \ref{fig:short1aaaasfgasgasdf2}, the weighted sum of the embeddings is calculated and passed to a final classification layer.    
Therefore, if a modality is missing during inference, the fusion module is robust, and the network \textit{still} yields an output prediction.

The model can be used as a one-size-fits-all solution to predict building age.      
We note      that the results are improved when we train \textit{two} different models. For the Full modality setting  (country ID: QCD, HUN), we use a SwinV2 Transformer in base size for embedding the street-view images and the VHR images. 
The Transformer has $87.9$~million parameters and expects inputs of $384\times 384$ pixels.   
It can take advantage of the high resolution of the street-view and VHR images. For the Top-View only setting (country ID: PNN, FMW), we use small SwinV2 Transformers ($49.7$~million parameters, $256\times 256$ pixels) for embedding VHR and S-2.   

\subsection{Results and Discussion}    

\noindent We  perform $5$-fold     stratified grouped cross validation. 
The dataset is split such that the image samples with the same city IDs  are grouped together. 
In addition, the split is stratified such that the class label distribution in each split is close to the overall class label distribution. 

We    first train    with all the available data. Then, we fine-tune the model using samples \textit{only}   from the countries QCD, FMW, and PNN, respectively.    
For predicting on HUN samples, we use the model fine-tuned on QCD samples.   
We use the cross-entropy loss function with class weights to counter class imbalance.   
Early stopping is applied with a patience of \textit{six} epochs.   
Training was performed on a NVIDIA A$100$ GPU with $40$~GB memory, and we note that fine-tuning the SwinV2 Transformers and training the fusion module takes no more than five epochs. We make predictions with the five pretrained models and choose the majority class. If there is a tie, we choose the building age that is more likely based on the distribution in the training data.



We     evaluate the model that is finally submitted to the development set. Here, the MAP is $0.7147$ on samples including all modalities, and $0.6013$ on samples including \textit{only} the top-view modalities. In Fig.~\ref{fig:confusionmatrix}, we show the confusion matrix for the samples that include         \textit{all} the modalities.      
The class labels are ordered from $0$ (oldest buildings) to $6$ (newest buildings).      
Here, we note that neighboring classes are often mistaken for each other (see label correlation in Sec. \ref{sec:labelcorrsec}). The network tends to overpredict the oldest building age category (i.e., the class $0$), likely because it is the most frequent class in the training dataset (see Fig.~\ref{fig:labels_dist}).

The code is available openly\footnote{\url{http://github.com/crlna16/ai4eo-map-your-city}}.      



\section{Baseline model: Age of Buildings for Sustainability (ABS)} \label{sec:baselinemo}

\begin{figure*}[t]                                   
  \centering                  
  \includegraphics[width=12.15cm]{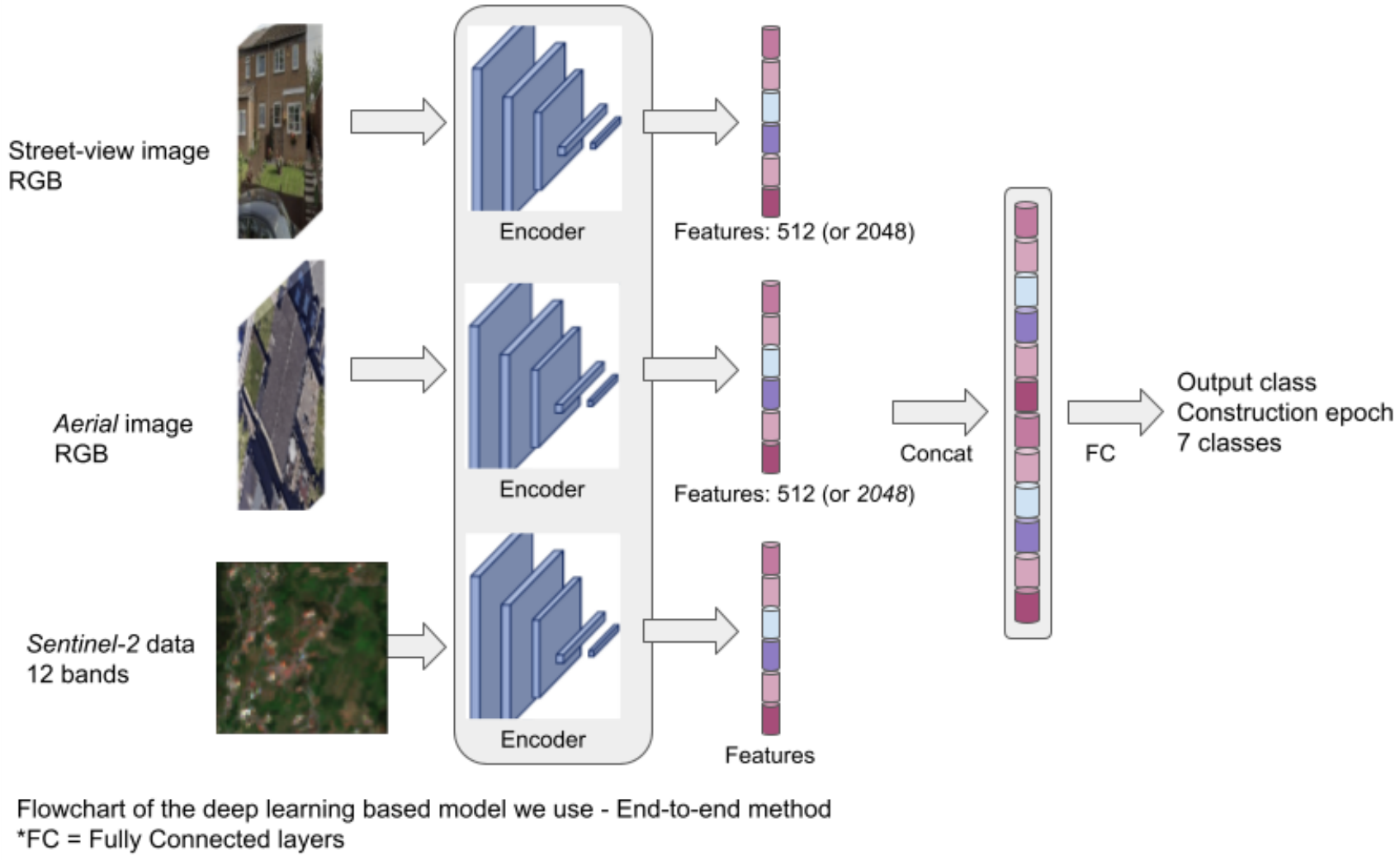}  
  \caption{Flowchart diagram of the ABS model, the baseline model of the challenge.}        
  \label{fig:shortfignew}               
\end{figure*}

\noindent This section introduces the baseline model of the MapYourCity Challenge, hereafter referred to as the ABS model (Age of Buildings for Sustainability). This model was included in the starter toolkit and made available to all challenge participants as a reference initial implementation. The proposed solution is outlined in Fig.~\ref{fig:shortfignew}.

\begin{figure}[t]                                      
  \centering \includegraphics[width=0.92\linewidth]{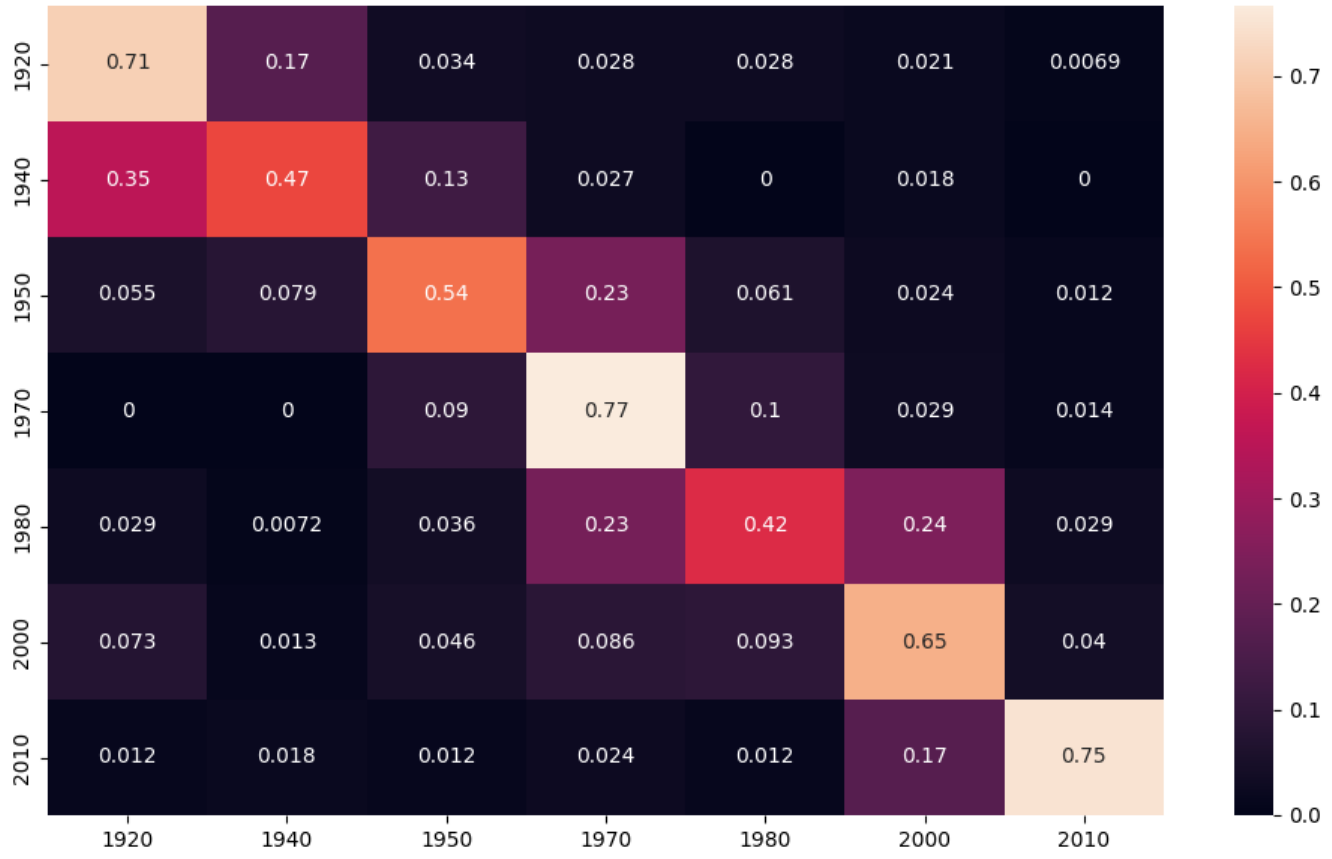} 
    \centering \includegraphics[width=0.92\linewidth]{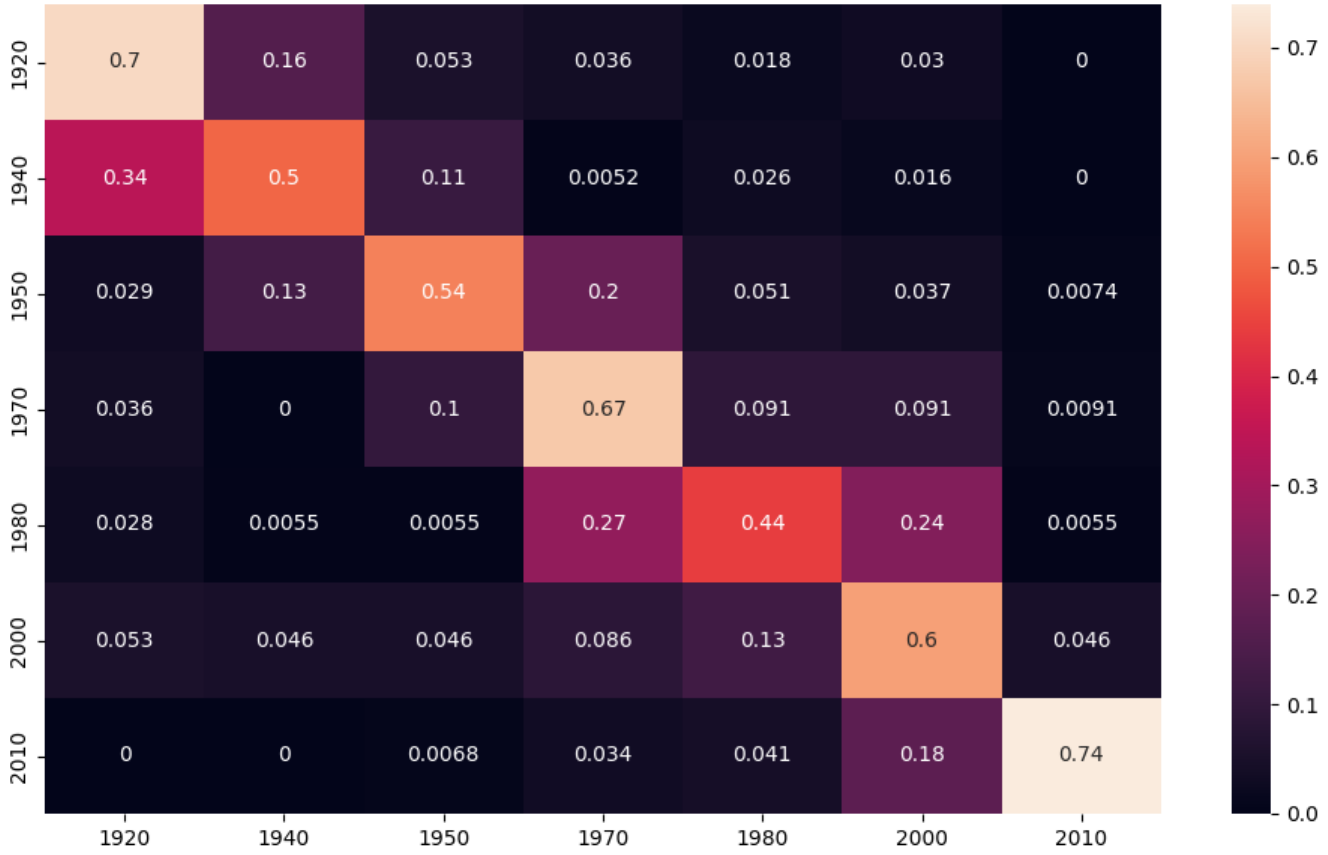}
  \caption{Classification confusion matrices evaluation and results for the baseline model ABS. Public leaderboard (Top) and Private leaderboard (Bottom).}        
  \label{fig:short1aaa}                      
\end{figure}





\subsection{Data Preparation} 
\noindent
For the ABS model, we apply data augmentation as well as random input dropout of street-view images.
More specifically, for data augmentation of street-view images, we employ horizontal flipping, Gaussian blur, and random picture hiding, i.e. insertion of \textit{all zeros}. For data augmentation of top-view VHR images, we use horizontal and vertical flipping, as well as Gaussian blur. In addition, for data augmentation of S-2, we employ Gaussian blur and random spectral band drop.


\subsection{Model}

\noindent
The ABS model is based on Transformers, and more specifically on the recently proposed model \textit{SegFormer} \cite{transf3, transf2}, which uses multi-scale features.      
The baseline model employs \textit{three} encoder networks, one for each modality, and performs late data fusion by concatenating the feature representations in the latent space.     




The ABS model is trained, validated, and evaluated on the multi-modal MapYourCity dataset. For the street-view and top-view VHR modalities, we employ SegFormer B5 as the encoder architecture. For S-2 imagery, we utilize an EO Foundation Model trained on unlabeled global S-2 data \cite{EOFM1, EOFM2}. This model was pre-trained using a geo-location prediction task and incorporates all spectral bands. Its architecture is based on a geo-aware U-Net design. The ABS model integrates these modality-specific models, each trained independently, to form a unified framework for estimating building age.



\begin{table*}[!tbp]    
\caption{ \centering {\color{black}Summary Table of the top-performing models. For the MPA metrics: the subscripts $_{f}$ and $_{t}$ refer to the MPA calculated with the Full modality and Top-View only settings, respectively. The subscripts $_{u}$ and $_{i}$ refer to the MPA calculated in the public and private leaderboards, respectively. ($\textbf{MPA}_{f-u}$) $|$ (Loss of MPA removing street-view modality $\textbf{MPA}_{t-u} - \textbf{MPA}_{f-u}$) $|$ Loss of MPA in between the public/private leaderboards for Full modality $\textbf{MPA}_{f-i} - \textbf{MPA}_{f-u}$) $|$ (Loss of MPA in between the public/private leaderboards for the Top-View setting $\textbf{MPA}_{t-i} - \textbf{MPA}_{t-u}$).}}    
\vspace{6.6pt}\centering       
\label{tab:maintablesumm} 
\vskip -0.0706in    
\vskip -0.0706in
\vskip 0in   
\begin{center}              
\begin{scriptsize}   
\begin{sc}         
\begin{tabular}          
{p{3.0cm} p{6.5cm} p{4.0cm}}
\toprule            
\midrule    
\textbf{Model} & \textbf{Main features}   & \textbf{MPA ($\%$)}   \\  
\midrule                        
\midrule         
First ranked team, Creamble (in Sec. \ref{sec:secfito})  & Label correlation modelling (see Fig. \ref{fig:figureprobablabel}, as well as Fig. \ref{fig:fignewfigdistlabels}), pretrained backbone EVA \cite{EVANEW}, model ensemble, hyper-parameter optimization Optuna, K-fold cross validation, Label smoothing with a smoothing factor of $0.3$, AdamW and CosineAnnealingWarmRestarts.   &   $76.0\%$	$|$ $-9.7\%$	$|$ $-2.7\%$ $|$	$-5.8\%$
    \\              
\midrule              
Second ranked team, TelePIX AI (in Sec. \ref{sec:sectop}) &  Pretrained backbone EVA-02 \cite{EVA2}, model ensemble, hyper-parameter optimization \textit{Optuna}, K-fold cross validation, Input dropout of street-view images.   & $75.8\%$	 $|$ $-5.9\%$ $|$	$-1.9\%$ $|$ $-11.9\%$
    \\     
\midrule                              
Third ranked team, xmb (in Sec. \ref{sec:sebe}) & Model ensemble, K-fold cross validation, Geometric fusion, Shared feature modelling, Input dropout of street-view images, Data augmentation CoarseDropout and GridDropout, Focal Loss with Label Smoothing (smoothing coeff.: $0.1$), AdamW and CosineAnnealingLR.     & $72.0\%$	$|$ $-3.0\%$ $|$	$-0.8\%$	$|$ $-10.9\%$
               \\        
\midrule                   
Fourth ranked team, The AI Buzzard (in Sec. \ref{sec:thibes}) & Model ensemble, pretrained backbone Swin Transformer SwinV2 encoders, K-fold cross validation, Late data fusion with weighted sum of the embeddings.   &   $71.8\%$ $|$	$-5.0\%$ $|$	$-0.8\%$ $|$	$-10.5\%$
       \\           
\midrule                  
\midrule 
Baseline model, ABS (in Sec. \ref{sec:baselinemo}) & Pretrained backbone SegFormer B5, Late data fusion, Input dropout of street-view images, Data augmentation such as Gaussian blur. & $61.6\%$ $|$ $-9.6\%$ $|$ $-1.7\%$ $|$ $-13.6\%$                  \\           
\midrule  
\bottomrule           
\end{tabular}     
\end{sc}           
\end{scriptsize}                
\end{center}                     
\vskip -0.1in         
\end{table*}

\subsection{Results and Discussion}   

\noindent Figure~\ref{fig:short1aaa} presents the classification confusion matrices for the ABS model, evaluated on the multi-modal MapYourCity dataset. The top panel corresponds to the public leaderboard, while the bottom panel shows results from the private leaderboard. These matrices reveal that the model frequently confuses adjacent classes, a pattern discussed already in Sec.~\ref{sec:labelcorrsec}. The overall MPA scores are $61.57$ for the public leaderboard and $59.86$ for the private leaderboard. The models submitted via the MapYourCity challenge (shown in Fig.~\ref{fig:short5}) demonstrate improved performance over the baseline ABS model, likely due to the use of model ensembling and/or hyperparameter optimization with Optuna. Additional enhancements such as label correlation, soft labeling, geometric data fusion, shared-specific feature modelling, Swin Transformer, and other key features further discussed in Table~\ref{tab:maintablesumm} and Sec.~\ref{sec:discdiscussion} also contribute to this improvement. These techniques collectively result in an approximate gain of $10\%$ in the MPA metric.

Additionally, since it does not employ model ensembling, the baseline ABS model has lower computational requirements compared to the models presented in Secs.~\ref{sec:secfito}–\ref{sec:thibes}. For instance, the ensemble described in Fig.~\ref{fig:short1aaaasfgasg1} and Sec.~\ref{sec:secfito} involves training $15$ separate models. {\color{black}Similarly, the model presented in Sec.~\ref{sec:sectop} also relies on an ensemble of $15$ models. This highlights the trade-off between performance and computational efficiency in model design.}
In this challenge, we evaluate the generalization capability of models to cities not included in the training set (see Sec.~\ref{sec:datatask} and Fig.~\ref{fig:short5}), ensuring strict separation between training and testing data to prevent leakage \cite{stratificationpaper}. Additionally, we assess model performance under modality constraints, specifically when the street-view data is unavailable.
For cities included in the training set, i.e., In-Distribution (ID), the ABS model achieves an MPA score of $64.61$. When the S-2 modality is excluded, performance drops to $60.11$, highlighting that despite its relatively low spatial resolution of 10 meters, S-2 still contributes meaningfully to the overall prediction. Furthermore, when both top-view modalities (VHR and S-2) are omitted, the MPA score decreases further to $58.72$. Comparing the ID MPA metric of $64.61$ with the MapYourCity challenge OoD MPA metric of $60.72$, the performance drop due to domain distribution shift appears reasonable, underscoring the inherent challenge of generalizing to previously unseen cities.

We also release our code for ABS for reproducibility\footnote{The GitHub repository is: \url{http://github.com/AI4EO/MapYourCity}.}.


\section{Overall discussion about the top-performing models}\label{sec:discdiscussion}

\noindent  
The \textit{main} characteristics shared by the models discussed in the preceding sections, where we note that the accuracy evaluation performance final results of the models is presented in Fig. \ref{fig:short5}, are as follows:      
\begin{itemize} 
    \item the use of a high-performing pretrained backbone, such as EVA \cite{EVANEW} or EVA-02 \cite{EVA2} (or Swin Transformer, e.g. SwinV2), to avoid training the model from scratch,
    \item model ensembling,   
    \item hyperparameter optimization, for example using Optuna \cite{optunapaper}. Here, we note that for example, Optuna is also used in the Earth Observation OoD detection model proposed in \cite{optuna2mainpaper}.
    \item modeling of label correlations between adjacent classes (per modality), based on the assumption that neighboring classes are more similar than non-adjacent ones, and
    \item a missing modality distribution alignment loss, combined with a \textit{shared} feature encoder.    
\end{itemize}

{\color{black}In addition to these core components, the models also incorporate several other key features, as summarized in Table~\ref{tab:maintablesumm}, such as K-fold cross-validation during training, geometric data fusion following the methodology proposed in \cite{FusionNew1}, and
the inclusion of a categorical country feature, represented as a one-hot encoded vector.



Several effective strategies were employed to enhance model robustness and performance. Notably, applying input dropout to street-view images during training proved valuable in addressing the missing modality problem at inference time, enhancing robustness when street-view inputs are unavailable. Additionally, the use of model ensembling and hyperparameter optimization also contributed to performance gains (i.e. see Table \ref{tab:maintablesumm}).


In addition, further important highlights are also using the \textit{Focal} Loss objective cost function to address the imbalance in the different classes in the multi-modal age of buildings dataset \cite{focallosspapercite} and performing data augmentation including, for example, \textit{CoarseDropout} and GridDropout.       
Moreover, geometric data fusion (i.e. as described in Sec.~\ref{sec:sebe} and   Fig.~\ref{fig:geometric_fusion}) and shared feature modelling (in Sec.~\ref{sec:sebe} and Fig.~\ref{fig:missing_modality}) improve model performance.

Furthermore, we also note that the methodology to first train with \textit{all} the available data and, then, fine-tune the model using samples \textit{only}   from specific countries (e.g., team The AI Buzzard, described in Sec. \ref{sec:thibes}) resembles \cite{ryd2025fine} and also EO Foundation Model (pre-)training and fine-tuning.         
Here, we also note that country-specific data for training were also used by Team TelePIX AI.

In Table \ref{tab:maintablesumm}, we also present and summarize the performance results of the different models.           
We observe for example that the winner, i.e. the first ranked team, Creamble, achieves high performance in the MPA evaluation metric. 
The model by Team Creamble was presented and discussed in Sec. \ref{sec:secfito}, and we also note that in the evaluation results, older classes in general are overpredicted by the models most likely because the number of data samples for the category `Before 1930' is larger than the number of data samples from the other classes (i.e. see Fig. \ref{fig:labels_dist}).

Regarding the robustness and sensitivity of the models to hyperparameter choices, we observe that the performance of the models is indeed sensitive to variations in these values. For instance, the top-performing model described in Sec.~\ref{sec:secfito} leverages Optuna to optimize the smoothing factor used in label smoothing.

Finally, regarding the computational resources that were used to train the models in Table \ref{tab:maintablesumm}, for \textit{most} of the models, the training was performed on a NVIDIA A$100$ GPU with $40$~GB memory.   
Here, we also note that the Team Creamble model uses two NVIDIA A6000 GPUs (i.e. see Sec. \ref{sec:mainmodels}).}

\section{Conclusion} \label{sec:conc}                      

\noindent In this work, we introduced MapYourCity, a novel multi-modal benchmark dataset designed to support real-world applications in EO generalization and urban analytics. By combining top-view VHR imagery, multi-spectral S-2 data, and co-localized street-view images, the dataset enables robust correlation learning and data fusion across modalities. The task is framed as a seven-class classification problem, targeting the construction epoch of buildings from 1900 to the present.
         
Through the MapYourCity Challenge, we demonstrated that estimating building age in previously unseen cities is both feasible and effective. Notably, models trained on the dataset can achieve good performance even when relying solely on top-view EO data, without access to street-view imagery, highlighting the scalability and global applicability of satellite-based approaches.

We presented and analyzed the top-performing models from the challenge, identifying key architectural and training strategies that contributed to their success, including label correlation modeling, ensemble learning, hyperparameter optimization, and late feature fusion techniques.

As future work, we plan to expand the MapYourCity dataset to include more cities globally and transition from discrete classification to continuous regression for construction year prediction. Additionally, we aim to incorporate explainability tools \cite{gradcam1, gradcampaper2, gradcampapermain3} such as Grad-CAM, saliency maps, and attention mechanisms to better understand model decisions and feature importance.

\section*{Acknowledgment}               

\noindent The      challenge was accomplished thanks to the contributions of many people.    
We want to thank all the team:  
Juan Pedro, Iacopo Modica, Dennis Albrecht, Annekatrien Debien, and everyone in AI4EO ESA \url{http://ai4eo.eu}.      

\ifCLASSOPTIONcaptionsoff 
  \newpage   
\fi

\bibliographystyle{IEEEtran}       
\bibliography{ref_all.bib}

\vspace{-38pt}        
\begin{IEEEbiography}[{\includegraphics[width=1in,height=1.in,clip,keepaspectratio]{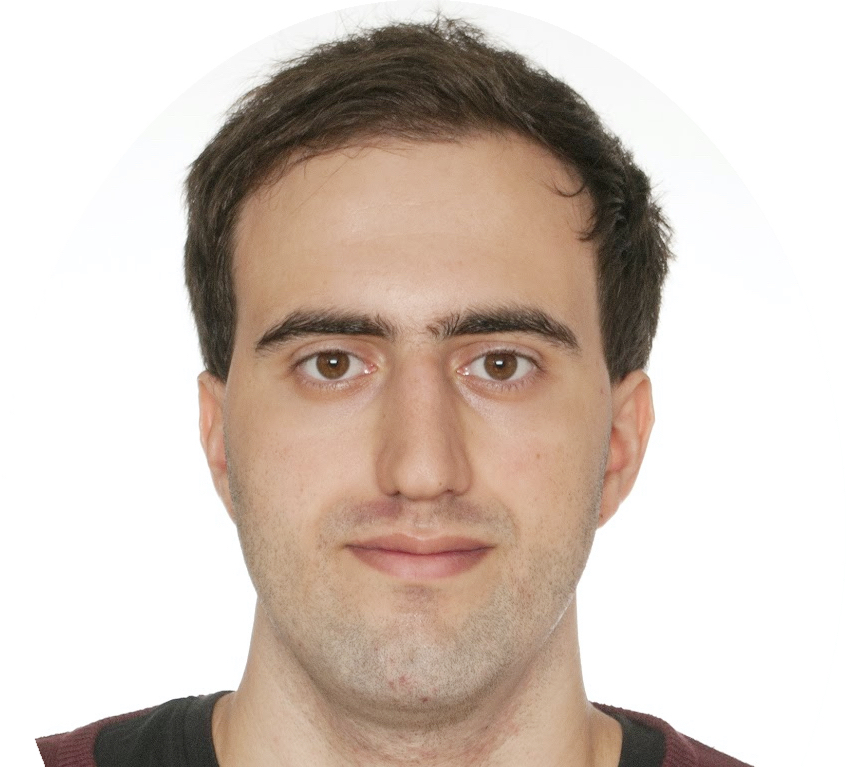}}]{Nikolaos Dionelis} \\
M.Eng. in electrical and electronic engineering from Imperial College London (ICL), UK and PhD degree in electrical engineering from ICL. Worked at the University of Edinburgh and the University Defence Research Collaboration (UDRC) in Signal Processing as a Postdoctoral Research Associate in Machine Learning for four years. Now, working at the European Space Agency (ESA), at the $\Phi$-lab.
\end{IEEEbiography}

\vspace{-42pt}     
\begin{IEEEbiography}  [{\includegraphics[width=1in,height=1.in,clip,keepaspectratio]{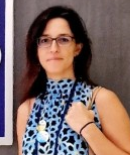}}]{Alessandra Feliciotti} \\
is a Project and Operations manager at MindEarth. She is also an Urban Scientist and with MindEarth, her tasks include urban and regional GIS analysis, webGIS applications and digital cartography. She also has a PhD degree from the University of Strathclyde, UK.
\end{IEEEbiography}

\vspace{-42pt}   
\begin{IEEEbiography}  [{\includegraphics[width=1in,height=1.in,clip,keepaspectratio]{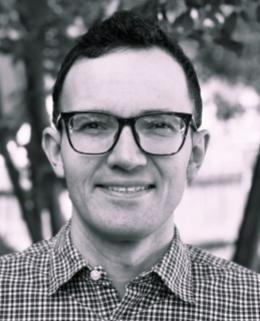}}]{Mattia Marconcini} \\
is a Research Data Scientist and Development Project Manager at the German Aerospace Center (DLR). He is an Engineer and Project Manager in the Smart Cities and Spatial Development team at the DLR Earth Observation Center (EOC) since March 2012. From August 2018, he has also been supporting MindEarth of which he has become partner in 2021. His research and work focus on the use of Earth Observation techniques to support sustainable urban development.
\end{IEEEbiography}

\vspace{-42pt}  
\begin{IEEEbiography}  [{\includegraphics[width=1in,height=1.in,clip,keepaspectratio]{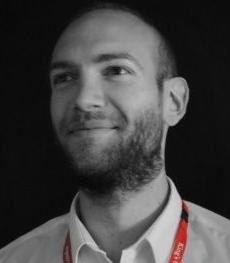}}]{Devis Peressutti} \\
is a Data/ML Scientist at Sinergise/ Planet. He develops machine learning solutions for remote sensing and earth observation applications. His research interests primarily lie in: machine learning and pattern recognition in remotely sensed imagery, image segmentation and registration, and prototyping and deploying large-scale ML algorithms to production.
\end{IEEEbiography}

\vspace{-42pt}     
\begin{IEEEbiography}  [{\includegraphics[width=1in,height=1.in,clip,keepaspectratio]{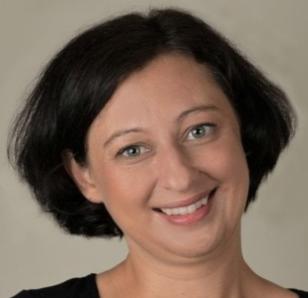}}]{Nika Oman Kadunc} \\
is a Software Engineer / Earth Observation Scientist at Sinergise/ Planet. She is a Data Scientist in the EO Research team discovering insights from earth observation data and turning creative ideas into working innovative solutions.
\end{IEEEbiography}

\vspace{-42pt} 
\begin{IEEEbiography}  [{\includegraphics[width=1in,height=1.in,clip,keepaspectratio]{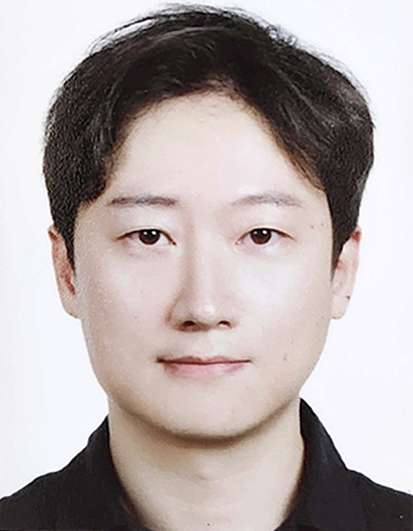}}]{JaeWan Park} \\
was born on March 26 , 1989, in Republic of Korea (South Korea).   
He received a Bachelor's degree in Economics from the Hankuk University of Foreign Studies and is currently pursuing a Master's degree in Data Science at the Sogang University. His research fields include detection tasks using satellite imagery, few-shot segmentation, and multi-modal approaches. He is currently working at the TelePIX AI Team, focusing on AI research and development using satellite imagery.
\end{IEEEbiography}

\vspace{-42pt} 
\begin{IEEEbiography}  [{\includegraphics[width=1in,height=1.in,clip,keepaspectratio]{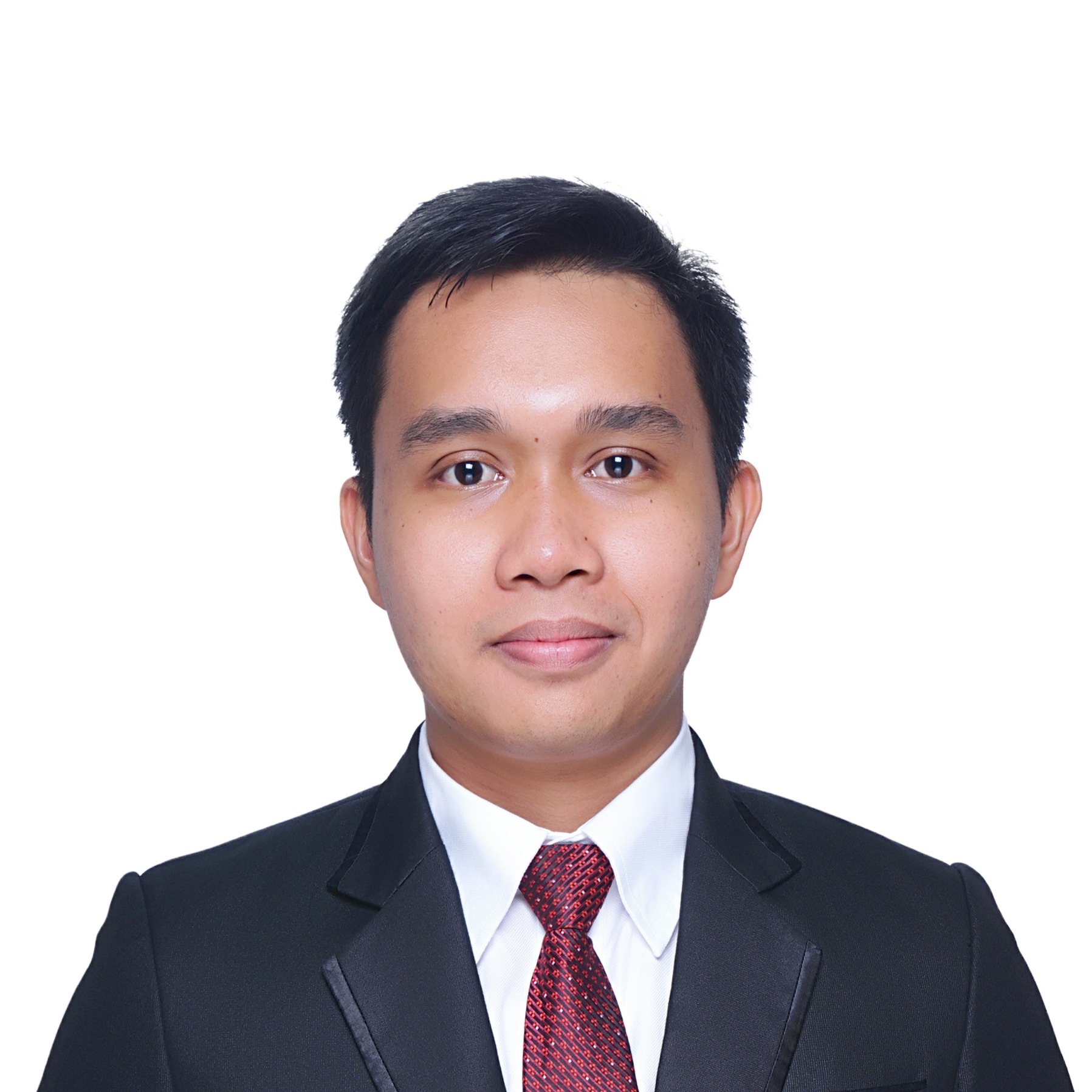}}]{Hagai Raja Sinulingga} \\
is an Indonesian computer scientist who earned his master's degree from Sejong University in 2023 and his bachelor's degree from Institut Teknologi Bandung in 2020. His expertise lies in detection tasks, few-shot segmentation, and foundational models. Currently, he is an artificial intelligence researcher and developer at the TelePIX AI Team, focusing on satellite imagery applications. Prior to joining TelePIX, he conducted research in anomaly detection at the Imaging and Intelligent Systems Lab, Sejong University.
\end{IEEEbiography}

\vspace{-28pt}  
\begin{IEEEbiography}   [{\includegraphics[width=1in,height=1.in,clip,keepaspectratio]{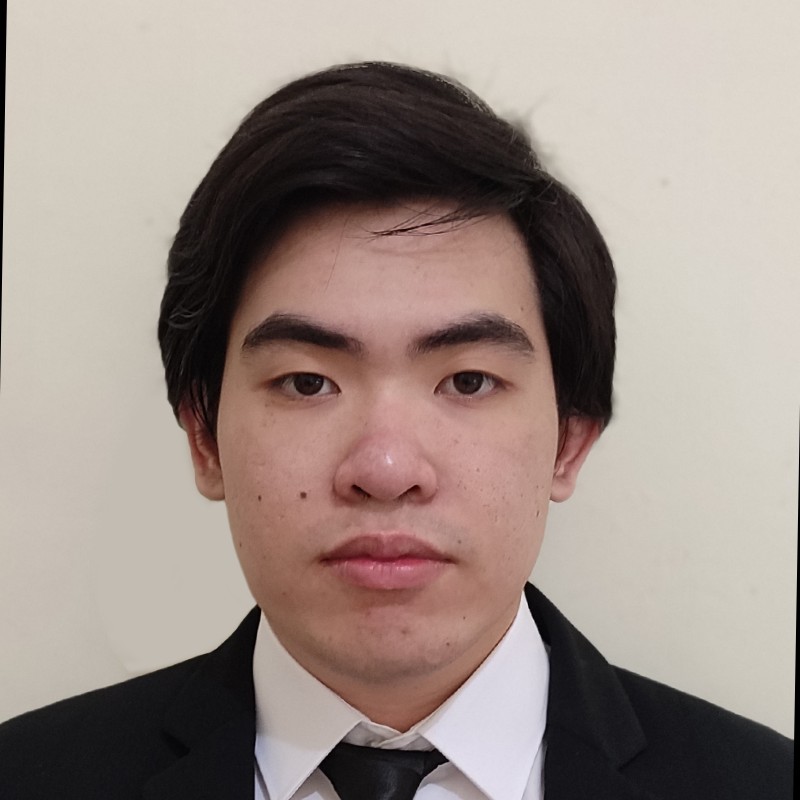}}]{Steve Andreas Immanuel } \\
was born in Indonesia. He completed his master’s degree at Sejong University in 2024, following his bachelor's from Institut Teknologi Bandung in 2021. His areas of expertise include vision-language models, few-shot segmentation, and image compression. He is currently an AI researcher and developer at the TelePIX AI Team, where he focuses on satellite imagery applications. Previously, he worked as a researcher at Vision Language Intelligence Lab, Sejong University.
\end{IEEEbiography}

\vspace{-40pt}   
\begin{IEEEbiography}  
[{\includegraphics[width=1in,height=1.in,clip,keepaspectratio]{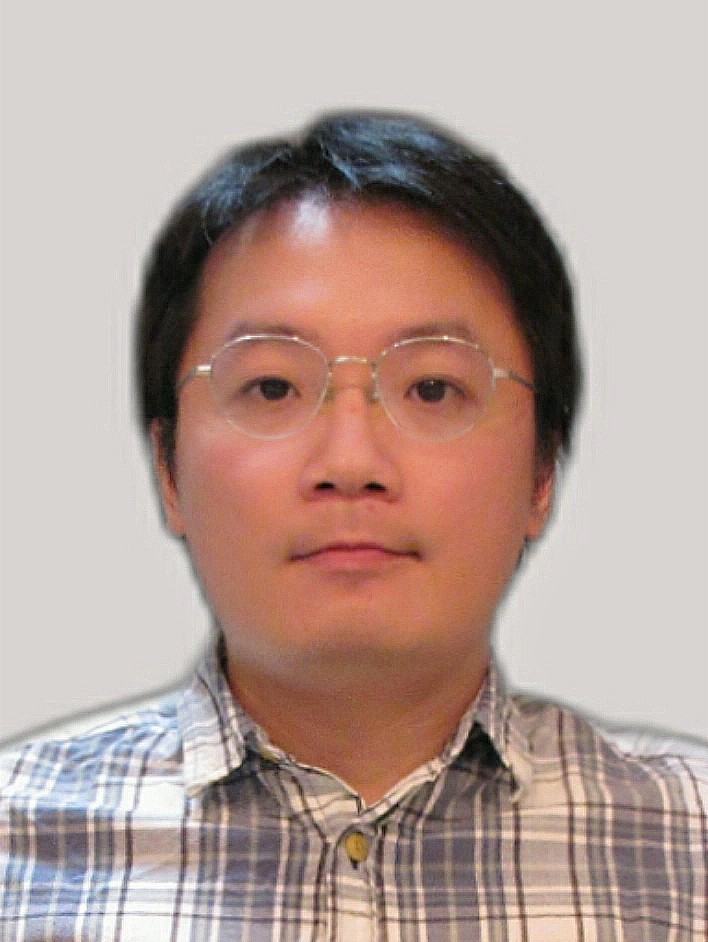}}]{Ba Tran} 
Ba Tran is a Computer Vision Software Engineer at Axelspace Corporation (Tokyo, Japan), specializing in satellite image construction for the GRUS satellite constellation and satellite image analysis. He graduated from the University of Tokyo in 2016, where he earned his Bachelor's degree in Information \& Communication Engineering from the Faculty of Engineering. He completed his Master's degree at the same university, in the Graduate School of Information Science \& Technology in 2018.
\end{IEEEbiography}

\vspace{-42pt}   
\begin{IEEEbiography}    [{\includegraphics[width=1in,height=1.in,clip,keepaspectratio]{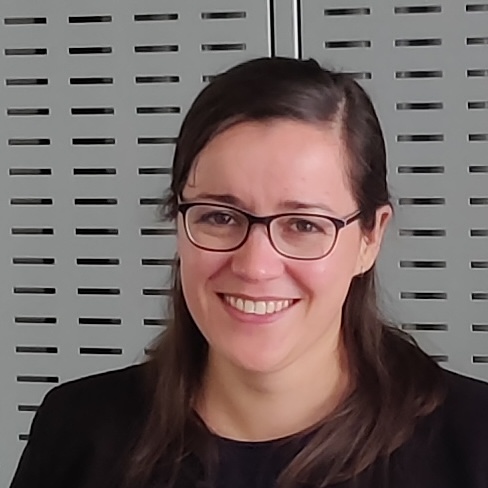}}]{Caroline Arnold} 
Caroline Arnold graduated in Physics from the University of Tübingen in 2015. She received a PhD in theoretical physics from the University of Hamburg in 2019. She joined Helmholtz AI and the German Climate Computing Center as an AI consultant in 2020. Her interests are deep learning for spatiotemporal data, remote sensing, and climate modeling.
\end{IEEEbiography}

\vspace{-42pt}            
\begin{IEEEbiography}[{\includegraphics[width=1in,height=1.in,clip,keepaspectratio]{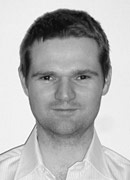}}]{Nicolas Longépé} \\ M.Eng. in electronics \& communication systems. M.S. degree in electronics at the National Institute for the Applied Sciences, France. PhD degree from Uni Rennes. Worked at EO Research Center of JAXA, as Japan Society for the Promotion of Science Fellow, and then as an invited researcher. Worked at CLS, France, as a research engineer in the Radar Application Division. Now, working at the European Space Agency (ESA), at the $\Phi$-lab.
\end{IEEEbiography}


\vfill

\end{document}